# How does the Performance of the Data-driven Traffic Flow Forecasting Models deteriorate with Increasing Forecasting Horizon? An Extensive Approach Considering Statistical, Machine Learning and Deep Learning Models

**Amanta Sherfenaz**
Graduate Research Assistant, Department of Civil Engineering
Bangladesh University of Engineering and Technology (BUET), Dhaka, Bangladesh, 1000
Tel: +880-2-55167100, Ext. 7225; Email: sherfenaz@gmail.com

**Nazmul Haque**
Lecturer, Accident Research Institute (ARI)
Bangladesh University of Engineering and Technology (BUET), Dhaka, Bangladesh, 1000
Tel: +8801673174808; Fax: +880-2-58613046; Email: nhaque@ari.buet.ac.bd

**Protiva Sadhukhan Prova**
Undergraduate Student, Department of Civil Engineering,
Bangladesh University of Engineering and Technology (BUET), Dhaka, Bangladesh, 1000
Tel: +8801752686927, Email: protivaprova24@gmail.com

**Md Asif Raihan***
Associate Professor, Accident Research Institute (ARI)
Bangladesh University of Engineering and Technology (BUET), Dhaka, Bangladesh, 1000
Tel: +8801911142802; Fax: +880-2-58610081; Email: raihan@ari.buet.ac.bd

**Md. Hadiuzzaman**
Professor, Department of Civil Engineering
Bangladesh University of Engineering and Technology (BUET), Dhaka, Bangladesh, 1000
Tel: +8801534951186; Fax: +880-2-58613046; Email: mhadiuzzaman@ce.buet.ac.bd

Word count: 6,227 words text + 2*250 (2 tables) = 6,727 words

*Corresponding Author

**ABSTRACT**

With rapid urbanization in recent decades, traffic congestion has intensified due to increased movement of people and goods. As planning shifts from demand-based to supply-oriented strategies, Intelligent Transportation Systems (ITS) have become essential for managing traffic within existing infrastructure. A core ITS function is traffic forecasting, enabling proactive measures like ramp metering, signal control, and dynamic routing through platforms such as Google Maps. This study assesses the performance of statistical, machine learning (ML), and deep learning (DL) models in forecasting traffic speed and flow using real-world data from California's Harbor Freeway, sourced from the Caltrans Performance Measurement System (PeMS). Each model was evaluated over 20 forecasting windows (up to 1 hour 40 minutes) using RMSE, MAE, and $R^2$ metrics. Results show ANFIS-GP performs best at early windows with RMSE of 0.038, MAE of 0.0276, and $R^2$ of 0.9983, while Bi-LSTM is more robust for medium-term prediction due to its capacity to model long-range temporal dependencies, achieving RMSE of 0.1863, MAE of 0.0833, and $R^2$ of 0.987 at a forecasting of 20. The degradation in model performance was quantified using logarithmic transformation, with slope values used to measure robustness. Among DL models, Bi-LSTM had the flattest slope (0.0454 RMSE, 0.0545 MAE for flow), whereas ANFIS-GP had 0.1058 for RMSE and 0.1037 for flow MAE. The study concludes by identifying hybrid models as a promising future direction.

## INTRODUCTION

Accurate traffic flow prediction is a critical component of modern Intelligent Transportation Systems (ITS), supporting proactive traffic management, real-time route optimization, and long-term infrastructure planning. With rapid urbanization and growing private vehicle ownership, traffic congestion has become a pressing global issue. According to the INRIX 2023 Global Traffic Scorecard, drivers in London lost an average of 156 hours annually in congestion, while Chicago and Paris reported over 138 hours of delay per driver, underscoring the severe economic and environmental costs of traffic inefficiencies (*1*). In this context, both short- and long-term traffic forecasting is essential for implementing adaptive signal timing, congestion pricing, and predictive routing. As connected and autonomous vehicle technologies evolve, the demand for accurate, high-resolution traffic prediction models becomes increasingly urgent to support real-time decisions and sustainable urban mobility.

Traffic flow and speed forecasting models can be classified into several categories starting with parametric models which have a predefined structure. Parametric models like ARIMA are commonly used for short-term forecasts due to their ability to model temporal autocorrelation in linear time series. However, they struggle with nonlinearity and spatial dependencies. Auto Regressive Moving Average (ARIMA) typically achieves $R^2$ between 0.6 and 0.75 for 15–30-minute forecasts, but performance drops sharply beyond one hour (*2*). Hybrid ARIMA–wavelet–neural models improve accuracy slightly but often ignore external factors like weather or incidents (*3*).

Other parametric models include filtering methods such as Kalman filters, which enable real-time sensor fusion and, when combined with connected vehicle and Bluetooth data, improve prediction accuracy by up to 11% over standalone methods (*2*). However, assumptions of linear Gaussian noise limit their robustness in complex traffic conditions.

ML models such as Random Forest, XGBoost (Extreme Gradient Boosting), Support Vector Machine (SVM), and k-Nearest Neighbor (KNN) can capture nonlinear traffic patterns. On PeMS data, Random Forest and XGBoost improve RMSE by 8–12% and reduce MAE by 10–15% compared to ARIMA. SVM and KNN are effective on smaller datasets but face scalability and computational challenges (*4*). ML models, however, show accuracy drops of up to 20% in RMSE during peak hours when traffic behavior is more chaotic (*2*), necessitating models that can learn complex spatiotemporal structures.

DL models, especially Recurrent Neural Networks (RNN) and variants like Long Short-Term Memory (LSTM) and Bi-LSTM, outperform simpler models by capturing nonlinear temporal dependencies. Bi-LSTM reduces RMSE by up to 25% over LSTM, with $R^2$ up to 0.85 for 15–30-minute forecasts (*5,6*). Hybrid CNN-LSTM models further improve spatial-temporal learning, with RMSE reductions of 10–15% for multi-hour forecasting (*7*).

Transformer-based models like Informer are good at long-range dependencies but are sometimes outperformed by simpler RNNs enhanced with periodic embeddings. For 30-day forecasts, RNNs with temporal embeddings reduced RMSE by up to 31.1% over Informer (*8*). DL models can overfit when data is limited or noisy. They also require heavy tuning and lack interpretability, which may hinder deployment in real-world systems.

Graph Convolutional Networks (GCNs), integrated with temporal models like LSTM (e.g., Temporal Graph Convolutional Network (T-GCN) and Diffusion Convolutional Neural Network (DCRNN)), effectively model road topology and spatiotemporal dependencies. These hybrid models reduce RMSE by up to 15% over CNN or RNN baselines on METR-LA and PeMS-Bay datasets, achieving $R^2$ above 0.9 (*9,8*). Attention mechanisms added to GCNs improve performance further, especially during congestion, with an additional 3–5% RMSE gain (*10*).

Forecasting horizon greatly impacts accuracy. Parametric models like ARIMA and Kalman filters perform reliably only for short windows under 20 minutes (*11*). Their RMSE may double for 60-minute forecasts (*2*). In contrast, ML and DL models maintain better accuracy over longer horizons.

XGBoost and Random Forest hold RMSE under 10 for up to 60 minutes, while LSTM models stay under 12 for 120 minutes. Hybrid CNN-LSTM and GCN-LSTM maintain RMSE below 13 even at 180 minutes. For multi-day forecasts, RNNs with temporal embeddings outperform Transformer-based models by up to 31.1% (*8*). Comparing models using RMSE, MAE, and $R^2$ across various horizons helps identify stable, high-performing models for operational deployment (*12,13*).

The above literature studies traffic forecasting in short-term and long-term and categorizes forecasts by time range (e.g. 15, 30 or 60 minutes ahead) and makes the prediction at a fixed prediction horizon. The concept of using sequential intervals into the future based on the data granularity (e.g. Every 5 minutes) have not yet been explored and a research gap lies in this area. Although some studies such as (*14,15,16*) mention multi-step forecasting, they primarily focus on deep learning architecture rather than exploring the performance of the models under the multi-step forecasting. This research gap is addressed in this study with the introduction of the concept of forecasting windows, which represent a multi-step ahead prediction interval to allow researchers to assess how the model performance changes across increasingly further time steps. The research objectives of this study can therefore be summarized as follows:

- To develop and evaluate a range of statistical, machine learning and deep learning models for short-term traffic flow and speed prediction using open-sourced high resolution loop detector data and comparing the performance of these models.
- To use the above models to predict traffic flow and speed across multiple foresting windows and quantize the change in performance in each model with increasing forecasting windows.

Thus, this study develops and compares multiple statistical, machine learning (ML), and deep learning (DL) models to forecast traffic flow on a selected road segment using real-world data from the California Department of Transportation's Performance Measurement System (PeMS). In addition to evaluating prediction accuracy, this study quantifies how the performance of each model deteriorates over increasing forecasting horizons, offering a novel framework for assessing model robustness that can guide future researchers in selecting appropriate models based on their temporal prediction needs.

## METHODOLOGY

In this study, several data-driven statistical, filtering methods, machine learning and deep learning models have been employed to predict the speed and flow across a roadway section. The performance of each model is compared to determine the best performing model. The models were then tuned to predict the speed and flow of vehicles across a roadway section for increasing forecasting windows to determine which models can withstand large forecasting windows. A brief description of the models used in this study are given in the following sub-section.

### Overview of the Models

As traffic flow and speed has continuously been evolving, researchers have attempted to model the flow and speed dynamics so that they can be predicted accurately both in the short-term and long term. This allows the alleviation of congestion, reducing emission and reducing travel times with an overall better traffic network efficiency (*2,15*). Researchers have developed various modeling techniques, starting with the simplest parametric mathematical statistics-based methods. However, these models were found to have low accuracy especially for highly complex non-linear traffic flow data and so combined with the availability of high-resolution large volumes of data from loop detectors and cameras, complex machine learning and eventually deep learning models were developed.

The simplest parametric models, such as linear regression attempts to establish linear relationship of the traffic flow and speed data. However, since traffic data is often complex and non-linear, linear regression was seen to perform poorly (*2,14*). Slightly more advanced models such Auto Regressive Moving Average (ARIMA), Kalman filters and Alpha Beta filters were seen to have better performance than linear regression the performance was still not accurate enough to real time application of the predicted traffic data. A study found ARIMA to have a Mean Absolute Percentage Error (MAPE) value of 18.95% for a 30-minute granularity data, Root Mean Squared Error (RMSE) of 189.3329 and Mean Absolute Error (MAE) of 100.3590 (*18*). An adaptive Kalman filter showed a significant margin of improvement of 11% in performance.

Machine learning models provide the advantage of the ability to extract complex non-linear characteristics of traffic data over these parametric models (*2,19*). Decision tree regression model, that uses multiple decision trees to predict values. An RMSE value of 84 and MAE value of 50 was obtained for a traffic flow prediction 60 minutes into the future. XGBoost was seen to perform better under the same dataset with an RMSE value of 66 and MAE value of 41 and Random Forest outperformed these models with RMSE 64 value of and MAE value of 38 (*15*). K-Nearest Neighbor, which predicts traffic flow based on feature extraction of nearest past value and the next value, was seen to perform more poorly than the other machine learning models. Another disadvantage of the KNN model is that since it uses a past value and the next value, it is only useful for training on static traffic data and cannot perform with real time data. Support Vector Machine (SVM) models are effective for traffic flow prediction because of its ability to eliminate noise using hyperplane that ignore small deviations and ability to capture non-linear data (*15*). Similarly, Artificial Neural Networks (ANN) can learn complex relationships of traffic flow data via a backpropagation mechanism. Each prediction error is fed back into the network to adjust weights, thus making it more suitable for high accuracy traffic forecasting. ANN was seen to have an RMSE of 163.97 and MAPE of 16.81% showing its better performance over ARIMA and other parametric models (*16*).

Deep learning models, which are extensions of machine learning models, have a higher capacity of capturing complex relationships of non-linear data. Long Short-Term Memory (LSTM) is able to retain past information in its memory cells making it ideal for modeling temporal flow data. LSTM was seen to have an RMSE of 96.3534, MAE of 55.8265, and WMAPE of 10.76% for 30-minute granularity (*6,13*) which is a significant improvement over the ANN and ARIMA models discussed above. Bi-LSTM is an improved structure of LSTM because it can process the data in both forward and backward direction (past and future value) which greatly improves the performance of the model (*5*). FC-LSTM is another improvement of the basic LSTM, that connect the hidden layers of the LSTM structure into one or more fully connected layers that allow it to refine complex temporal patterns to map the learned features to predict traffic flow (*15*). Another deep learning model commonly used is the Fated Recurrent Unit (GRU) that uses a simplified gating mechanism similar to LSTM, but the simplification reduces computational complexity making it more suitable for traffic flow and speed prediction. Adaptive Neuro Fuzzy Inference Systems (ANFIS) are a popular deep learning technique foe prediction of sequential data. ANFIS combines neural networks with fuzzy logic, using a layered structure to learn fuzzy rules from data, making it effective for modeling nonlinear relationships in traffic flow. The Grid Partitioning (GP) method generates fuzzy rules by dividing the input space into uniform grids, which can lead to high model complexity with many inputs but ensures complete rule coverage. In contrast, Subtractive Clustering (SC) and Fuzzy C-Means Clustering (FCMC) create more compact rule sets by identifying cluster centers based on data density or membership degrees, making them more efficient and better suited for handling uncertainty and overlapping traffic states (*16,17*).

Other deep learning models that offer superior performance of traffic flow prediction with advanced model architecture are Convolutional Neural Networks (CNN) that are able to extract spatial features of flow by treating the traffic network as a grid data. Sequence to sequence encoders learn hierarchical features from neighboring points achieving low MAE and RMSE values (*7*).

The diversity of the models overviewed above, ranging from parametric to machine learning to deep learning models are a proof of the evolving complexity of traffic flow and speed prediction tasks. Each model has their unique strengths with ML and DL techniques having a complex acritude that make it more suitable for modeling the complex nature of traffic flow. These models are used to predict traffic flow in this study and the hyperparameters used for these models are discussed in the next section.

**Model Performance**

In this study, performance measures of training root mean squared error (training RMSE), testing root mean squared error (testing RMSE), training mean absolute error (training MAE), testing mean absolute error (testing MAE) and overall goodness of fit ($r^2$) of the model are evaluated for each model and for each of the 20-forecasting window. The RMSE, defined as the standard deviation of the predicted values of flow and speed from the actual value and is calculated as shown in **Equation 1** where $y_i$ are the true values and $\widehat{y_i}$ are the predicted values.

$$RMSE = \sqrt{\frac{1}{n}\sum_{i=1}^{n}(y_i - \widehat{y_i})^2} \tag{1}$$

The mean absolute error (MAE) is defined as the average of the absolute value of the differences between the true values ($y_i$) and the predicted values ($\widehat{y_i}$). The MAE is calculated in this study as shown in **Equation 2** where $y_i$ are the true values and $\widehat{y_i}$ are the predicted values.

$$MAE = \frac{1}{n}\sum_{i=1}^{n}|y_i - \widehat{y_i}| \tag{2}$$

The goodness of fit value $R^2$, which is a measure of how well the models fit the given dataset is calculated as shown in **Equation 3**, where $y_i$ are the true values, $\widehat{y_i}$ are the predicted values and $\bar{y}$ is the mean of the $y_i$ data.

$$R^2 = 1 - \frac{\sum_{i=1}^{n}(y_i - \widehat{y_i})^2}{\sum_{i=1}^{n}(y_i - \overline{y_i})^2} \tag{3}$$

**Description of Dataset**

Field data of a roadway segment has been obtained from the California Department of Transportation (Caltrans) Performance Measurement System (PeMS) database. The speed and flow data has been downloaded from the Caltrans website (*20*). The MVDS 763663 (S of 91) station situated at 110 Harbor Freeway in California to the South of the intersection of the Freeway and Gardena freeway (**Figure 1a**) with the station at coordinates (33.870044, -118.28482). The road segment where the station is placed has a width of 60 ft and consists of five lanes with a design speed limit of 70 mph. The speed and flow data of five weeks from 01/01/2024 to 04/01/2024 has been downloaded which has a temporal granularity of 5 minutes. The aggregate data of all five lanes have been used as the input for the prediction models as a sequential input and was split into 80% for training, 10% for validation and 10% for testing for the machine learning and deep learning models.

The downloaded data contains the speed and flow of each of the five lanes in a columnar format and also aggregated speed and flow as shown in column L and M (**Figure 1b**). This aggregate data is imported into MATLAB environment and the data is then trained into the statistical, machine learning and deep learning models in a Windows 11, 64-bit machine with the ‘Deep Learning Toolbox’, ‘Statistics and Machine Learning Toolbox’ and ‘Fuzzy Logic Toolbox’ along with other basic toolboxes installed.

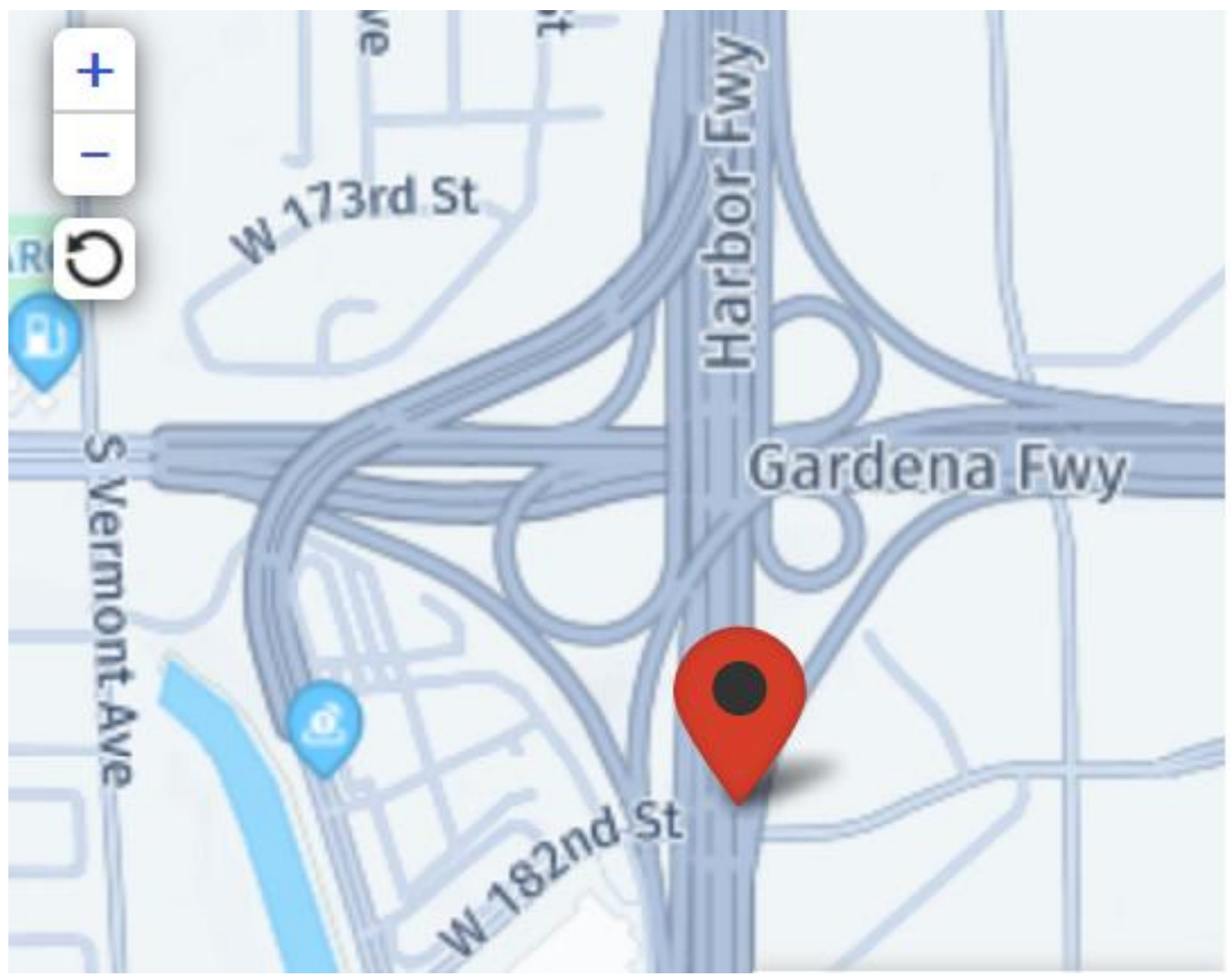

(a)

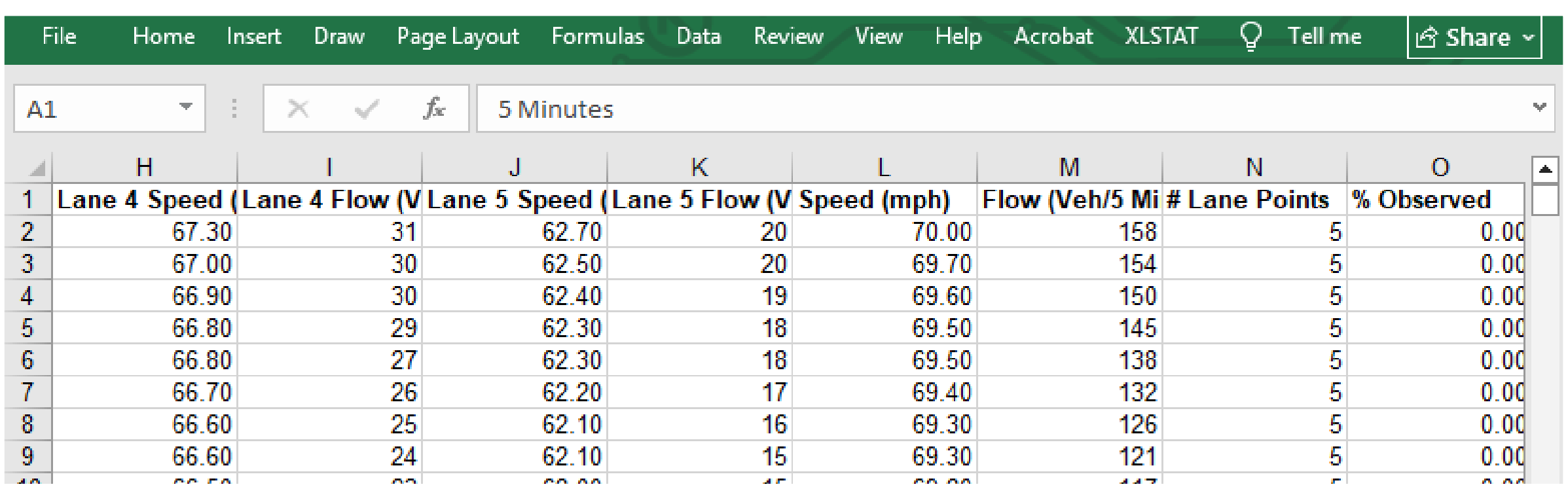

| | H | I | J | K | L | M | N | O |
|---|---|---|---|---|---|---|---|---|
| 1 | Lane 4 Speed ( | Lane 4 Flow (V | Lane 5 Speed ( | Lane 5 Flow (V | Speed (mph) | Flow (Veh/5 Mi | # Lane Points | % Observed |
| 2 | 67.30 | 31 | 62.70 | 20 | 70.00 | 158 | 5 | 0.00 |
| 3 | 67.00 | 30 | 62.50 | 20 | 69.70 | 154 | 5 | 0.00 |
| 4 | 66.90 | 30 | 62.40 | 19 | 69.60 | 150 | 5 | 0.00 |
| 5 | 66.80 | 29 | 62.30 | 18 | 69.50 | 145 | 5 | 0.00 |
| 6 | 66.80 | 27 | 62.30 | 18 | 69.50 | 138 | 5 | 0.00 |
| 7 | 66.70 | 26 | 62.20 | 17 | 69.40 | 132 | 5 | 0.00 |
| 8 | 66.60 | 25 | 62.10 | 16 | 69.30 | 126 | 5 | 0.00 |
| 9 | 66.60 | 24 | 62.10 | 15 | 69.30 | 121 | 5 | 0.00 |

(b)

**FIGURE 1**. Details of data collection (a) Location of MVDS 763663 station (b) Sample of collected data

## MODEL HYPERPARAMETER SETTINGS

The models used in this study can be broadly classified into parametric, non-parametric model and deep learning methods (*12*) (**Table 1**). Parametric models used in this study are Linear Regression, Auto Regressive Moving Average (ARIMA), Kalman Filter and Alpha Beta Filter. Linear regression is the simplest form of statistical modeling with only two parameters that are calibrated with the data and used to predict traffic flow and speed. However, it cannot fully capture the temporal variation of the data. Auto Regressive Moving Average (ARIMA) overcomes this limitation by using three hyperparameters of an Auto Regressive (AR) term, differencing the data to eliminate trends and using a moving average (MA) to consider the past errors (*18*). In this study, 2 AR terms are used so that the last two values of the speed and flow sequences are used to predict the next value, and the degree of differencing is set to 1 so that the data is treated only once to remove trends and 2 errors of the past two iterations are used to predict the next value.

The prediction of traffic flow and speed often require real time forecasting and may use noisy field data which cannot be done using linear regression and ARIMA because these can only handle static recorded data. Filtering methods such as Kalman filter and Alpha Beta filter can model in real time by continuously updating model coefficients in real time (*2*). These can also handle noisy data as they effectively filter out noisy data using filtering coefficients. In this study, training hyperparameters Q = 0.001 which determines model uncertainty and R = 0.01 which determines data uncertainly are used filter out noisy data and predict traffic flow and speed using Kalman filter. Hyperparameters $\alpha = 0.85$ and $\beta = 0.005$ are used for Alpha Beta filtering to, also, filter out noisy data and dynamically predict traffic flow and speed sequences.

Parametric models discussed above assume linearity of data and therefore cannot handle the complex traffic flow patterns in peak hours and during incidents (*18*). These models require a large dataset in order to achieve sufficiently good performance. Machine learning techniques, on the other hand, can work on small datasets by learning the patterns in traffic flow and speed which can non-linear, piecewise and having interacting effects within the data since they do not have a predetermined structure. In this study, machine learning model, decision trees, are used with hyperparameter 'SplitCriterion' set to 'mse' so that the trees are split such that the mean squared error (MSE) values are minimized. The hyperparameters used for the XGBoost model using 'LSBoost' method are 'MaxNumSplits' = 10 so that a decision tree can split into maximum 10 to prevent overfitting of the data and 'NumLearningCycles' = 100 for iterating the trees a hundred times (**Table 1**).

The hyperparameters used in the Random Forest model are a window size of 10 to use the past 10 values to predict the next value of the sequence and 100 decision trees to train the model by regression and using a built-in validation 'Out-of-bag (OOB)' method to generalize the data. For the k-Nearest Neighbor model, the number of nearest neighbors used is 5, meaning that 5 nearest data points are used to prediction and the number of past values (sequence length) used to predict is also 5. This ensures avoidance of overfitting or underfitting and that enough temporal context of the sequences are captured to adequately reflect the data. The hyperparameters used for the Support Vector Machine model are k =1 for predicting one step ahead at a time and the radial basis function (rbf) kernel is used making the model suitable for prediction of non-linear and complex traffic flow and speed data. Another machine learning model used in this study is the Artificial Neural Network (ANN) with the hyperparameters of using 10 neurons in the hidden layer to control the model capacity and optimizing the network using a Levenberg-Marquardt method.

The deep learning methods used in this study, are extensions of the machine learning model but with better capacity of handling large datasets to extract patterns in data. In the Convulutional Neural Network, several hyperparamters are used such as the lag size (k) with a value of 5 that uses the past 5 values to predict the next value are used in this study. These values are used to form a 5×1 "image" for the 'imageInputLayer' and two 'convolution2dLayer' with 16 and 32 channels that uses 3 adjacent data to learn local temporal patterns. A Rectified Linear Unit (ReLu) architecture is used in this study for the CNN model with 64 neurons in the fully connected layer with a 'adam' optimizer and training every 100 iterations. Similarly, for the sequence-to-sequence Encoder model, a window size of 10 is used and 100 neurons in the lstm hidden layer with an 'adam' optimizer. For each training step, 32 sequences are trained together (MiniBatchSize) and the total dataset is trained in 100 cycles (epochs). For the LSTM, Bi-LSTM and FC-LSTM models, 100 neurons in the hidden layers are used for LSTM and FC-LSTM whereas 128 hidden neurons are used for Bi-LSTM. All three models use the 'adam' optimizer and data is shuffled 'every-epoch' for LSTM and Bi-LSTM but 'never' for FC-LSTM. The GRU model used in this study also has a similar structure with 128 neurons in the hidden layer and an 'adam' optimizer that passes over the dataset 200 times (epochs), shuffling the data after every training step. In the three variations of the ANFIS model used in this study, the 'genfisOptions' hyperparameters are set to 'GridPartition', 'SubtractiveClustering' and 'FCMClustering' respectively. For the ANFIS-GP model, 3 fuzzy membership generalized bell shaped functions are used for each input (gbellmf).

The ANFIS-SC model uses three clusters of influence radius 0.5, 0.25 and 3 to use smaller clusters to perform better under the dataset. The ANFIS-FCMC also works on three fuzzy clusters and a FIS-type 'sugeno'. All three models uses an InitialFIS of (fis_X, fis_Y) of the Fuzzy Inference Systems created by the three FIS structures (GP, SC or FCMC). Using the models discussed above, the aggregate flow and speed data were split into training, testing and validation parts. The models were used to predict the traffic flow and speed individually for 20 forecasting windows which means the prediction was done for the next 100 minutes.

**TABLE 1**. Hyperparameters of the models

| Model classification | Model | Hyperparameter | Value |
|---|---|---|---|
| Parametric | Linear Regression | - | - |
| | ARIMA | Auto Regressive term (AR) | 2 |
| | | Degree of differencing (d) | 1 |
| | | Moving Average (MA) order | 2 |
| | Kalman Filter | Q | 0.001 |
| | | R | 0.01 |
| | Alpha Beta Filter | α | 0.85 |
| | | β | 0.005 |
| Machine Learning | Decision Tree Regression | SplitCriterion | mse |
| | XGBoost | MaxNumSplits | 10 |
| | | Method | LSBoost |
| | | NumLearningCycles | 100 |
| | Random Forest | windowSize | 10 |
| | | numTrees | 100 |
| | | Method | Regression |
| | | OOBPrediction | On |
| | K-Nearest Neighbour (KNN) | 'k' | 5 |
| | | sequence_length | 5 |
| | Support Vector Machines (SVM) | 'k' | 1 |
| | | KernelFunction | rbf |
| | Artificial Neural Network (ANN) | Hidden neurons | 10 |
| | | Training algorithm | trainlm |
| Deep learning | Convolutional Neural Network (CNN) | k | 5 |
| | | imageInputLayer | [5 1 1] |
| | | convolution2dLayer (1st) | 16 |
| | | convolution2dLayer (2nd) | 32 |
| | | activation | relu |
| | | fullyConnectedLayer | 64 |
| | | optimizer | adam |
| | | epoch | 100 |
| | Encoders | windowSize | 10 |
| | | numHiddenUnits | 100 |
| | | optimizer | adam |
| | | MiniBatchSize | 32 |
| | | epochs | 100 |
| | LSTM | lstmLayer | 100 |
| | | Optimizer | adam |
| | | MaxEpochs | 100 |
| | | Shuffle | every-epoch |
| | Bi-LSTM | num_hidden_layer | 128 |
| | | MaxEpochs | 200 |
| | | Optimizer | adam |
| | FC-LSTM | numHiddenUnits | 100 |
| | | MaxEpochs | 100 |
| | | Optimizer | adam |
| | GRU | num_hidden_layer | 128 |
| | | Optimizer | adam |
| | | MaxEpochs | 200 |
| | | Shuffle | every-epoch |
| | ANFIS-GP | genfisOptions | 'GridPartition' |
| | | NumMembershipFunctions | 3 |
| | | InputMembershipFunctionType | gbellmf |
| | | InitialFIS | fis_X, fis_Y |
| | ANFIS-SC | genfisOptions | 'SubtractiveClustering' |
| | | ClusterInfluenceRange | [0.5 0.25 0.3] |
| | | InitialFIS | fis_X, fis_Y |
| | ANFIS-FCMC | genfisOptions | 'FCMClustering' |
| | | NumClusters | 3 |
| | | FISType | sugeno |
| | | InitialFIS | fis_X, fis_Y |

## RESULTS AND DISCUSSION

### Performance of the Models

The performance metrics of the models for forecasting windows 1, 10 and 20 is shown in **Figure 2**. As found in previous literature, the training RMSE and training MAE of the statistical and parametric models such as linear regression, ARIMA as well as filtering methods of Kalman filter and alpha beta filter are the highest (*1-4*). The linear regression and ARIMA models are seen to have an RMSE of 5.01 and 6.88 for speed prediction and 159.20 and 381.99 for flow at forecasting window 1. The $R^2$ value of the linear regression model is also very low at 0 for linear regression and -4.75 for ARIMA prediction of flow indicating very poor fit as is consistent with the findings of previous literature (*19*). Tree based and classical machine learning models such as XGBoost, Random Forest, KNN and SVM performed slightly better with RMSE values of 1.33, 1.09, 1.60 and 6.79 respectively for flow prediction. From Fig 2, it is seen that among the machine learning methods, the Alpha Beta Filter is the most effective at predicting the flow with the lowest RMSE in the group of 0.37. However, for speed prediction, AB filter has a high RMSE of 8.16 with the most accurate model from parametric and machine learning model to be XGBoost with an RMSE of 0.07 and an $R^2$ value of 0.999. This is because AB filter assumes constant velocity and therefore ignoring speed fluctuations resulting in its poor performance with speed data but suitable for flow prediction because of the less erratic flow data. The performance of XGBoost for speed prediction is better than for flow prediction because it uses a tree-based architecture that can adapt to sudden fluctuation in data, making it ideal for speed prediction because drivers often hit brakes at any sudden stimulus.

Among all the models combined, the deep learning models show the best performance for both speed and flow prediction (**Figure 2**). ANFIS-GP is seen to perform the best for both speed and flow with RMSE at 0.058 and 0.038, MAE of 0.0365 and 0.0276 and $R^2$ of 0.99 for both models respectively at forecasting window 1. This is because ANFIS-GP architecture has the ability to leverage the fuzzy inputs into a grid structure that captures complex, multidimensional patterns in traffic data. Unlike ANFIS-SC which may underfit for low data density, ANFIS-GP guarantees full coverage of the dataset. Hence, it is seen to perform the best among the other models.

In **Figure 2**, at higher forecasting window of 10 and 20, it is seen that the best performing model is now Bi-LSTM for both speed and flow with RMSE of 0.1029 and 0.1394, MAE of 0.0535 and 0.0559 and $R^2$ value of 0.99 for both models at forecasting window of 10. Similarly, the RMSE of 0.0.1423 and 0.0.2829, MAE of 0.0763 and 0.0883 and $R^2$ value of 0.99 and 0.91 for speed and flow respectively is obtained with Bi-LSTM at forecasting window of 20. This shows that at larger forecasting windows, the accuracy of ANFIS-GP decreases at a greater rate than for Bi-LSTM making it more robust for predicting traffic speed and flow further into the future. This is because ANFIS-GP uses immediate past data for predicting the next and at further forecasting windows the errors of the past values become

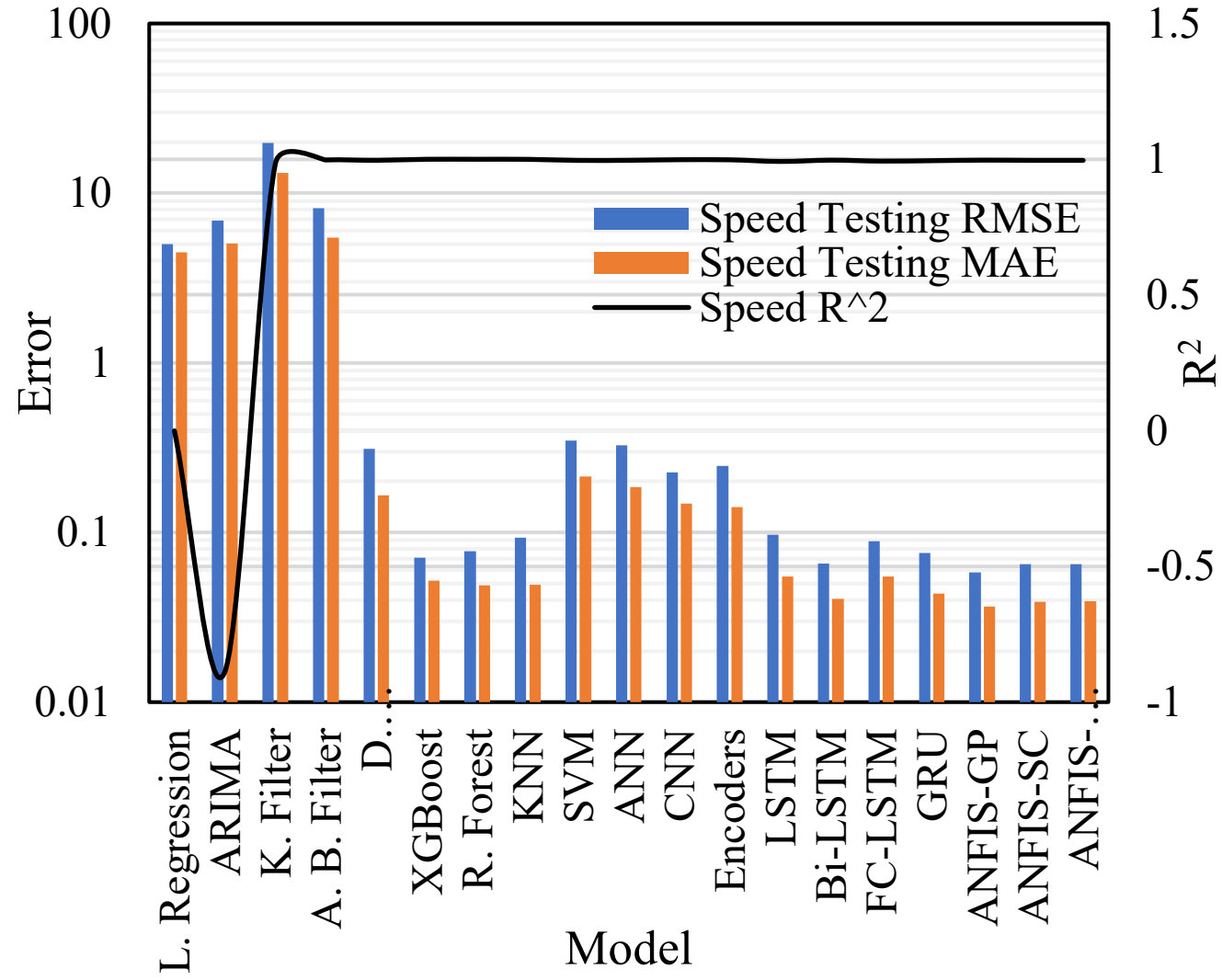

(a) Performance of models for speed at forecasting window 1

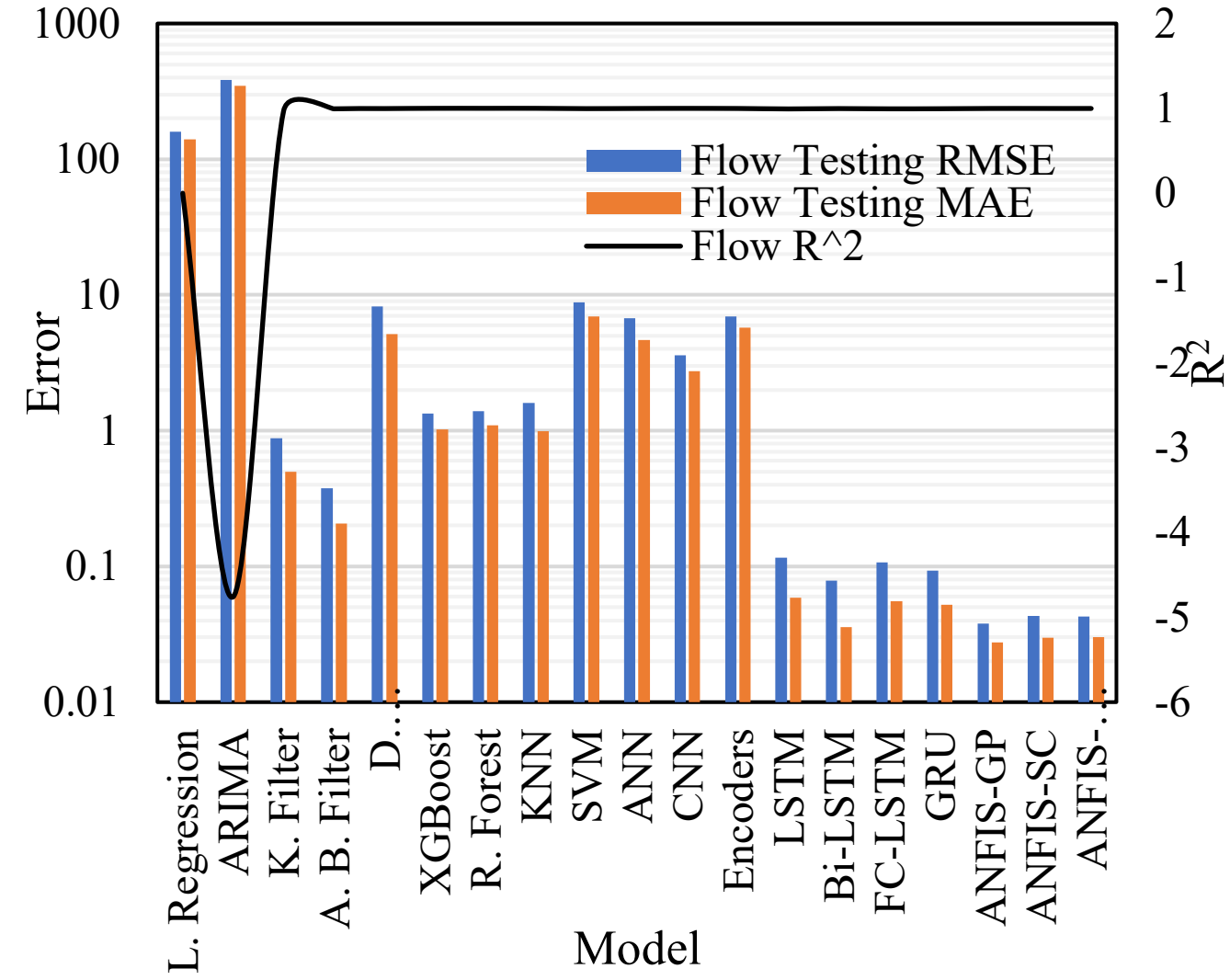

(b) Performance of models for flow at forecasting window 1

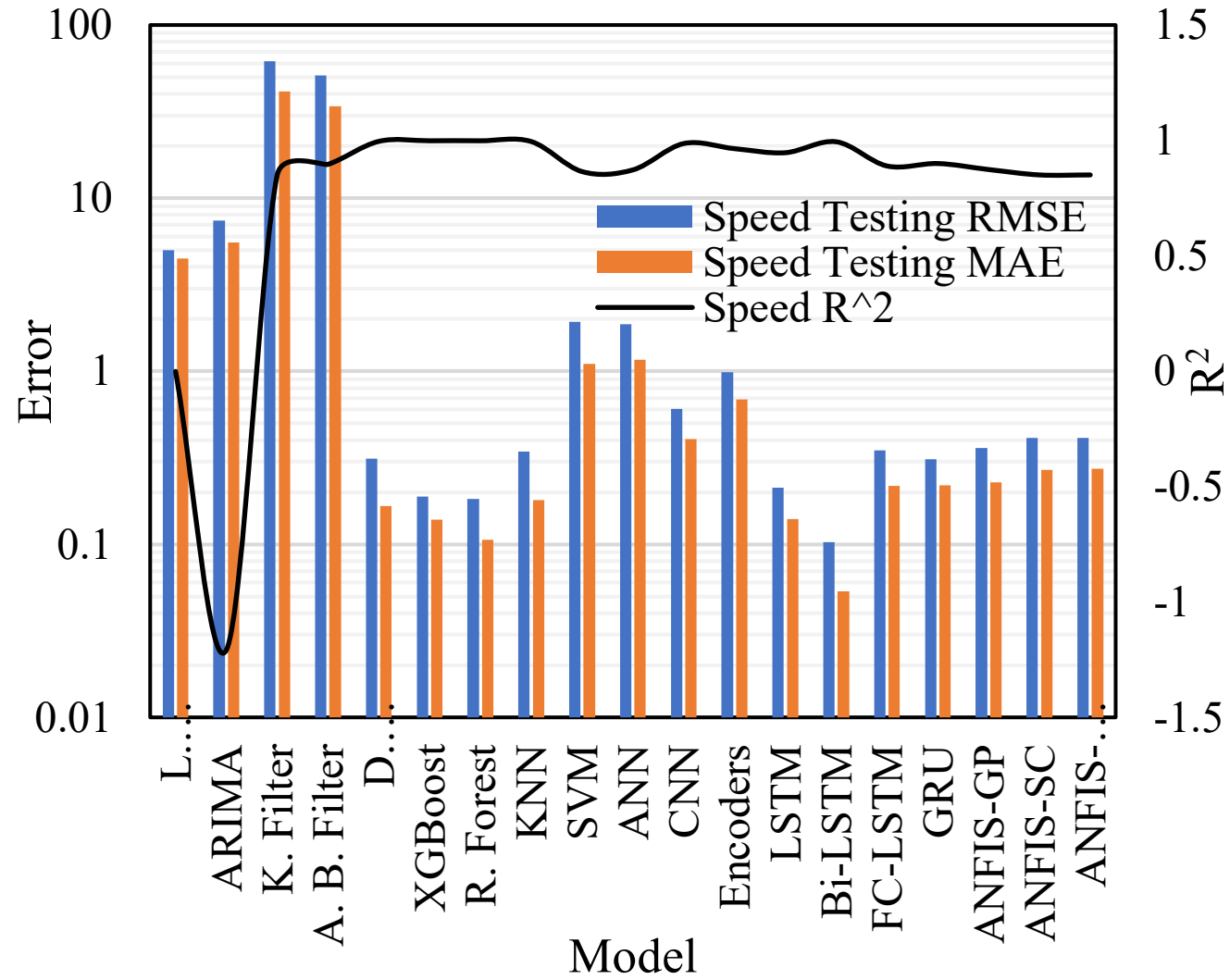

(c) Performance of models for speed at forecasting window 10

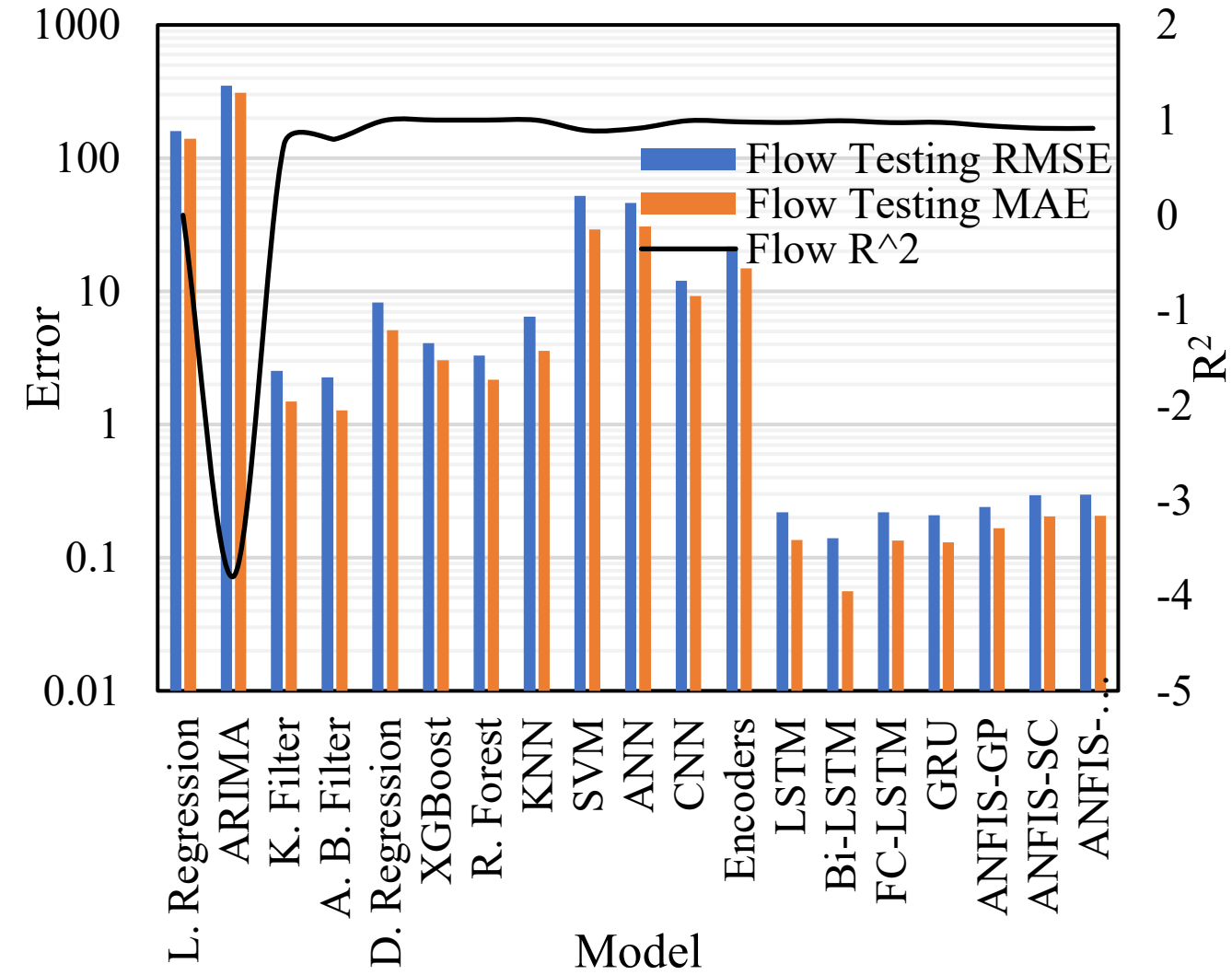

(d) Performance of models for flow at forecasting window 10

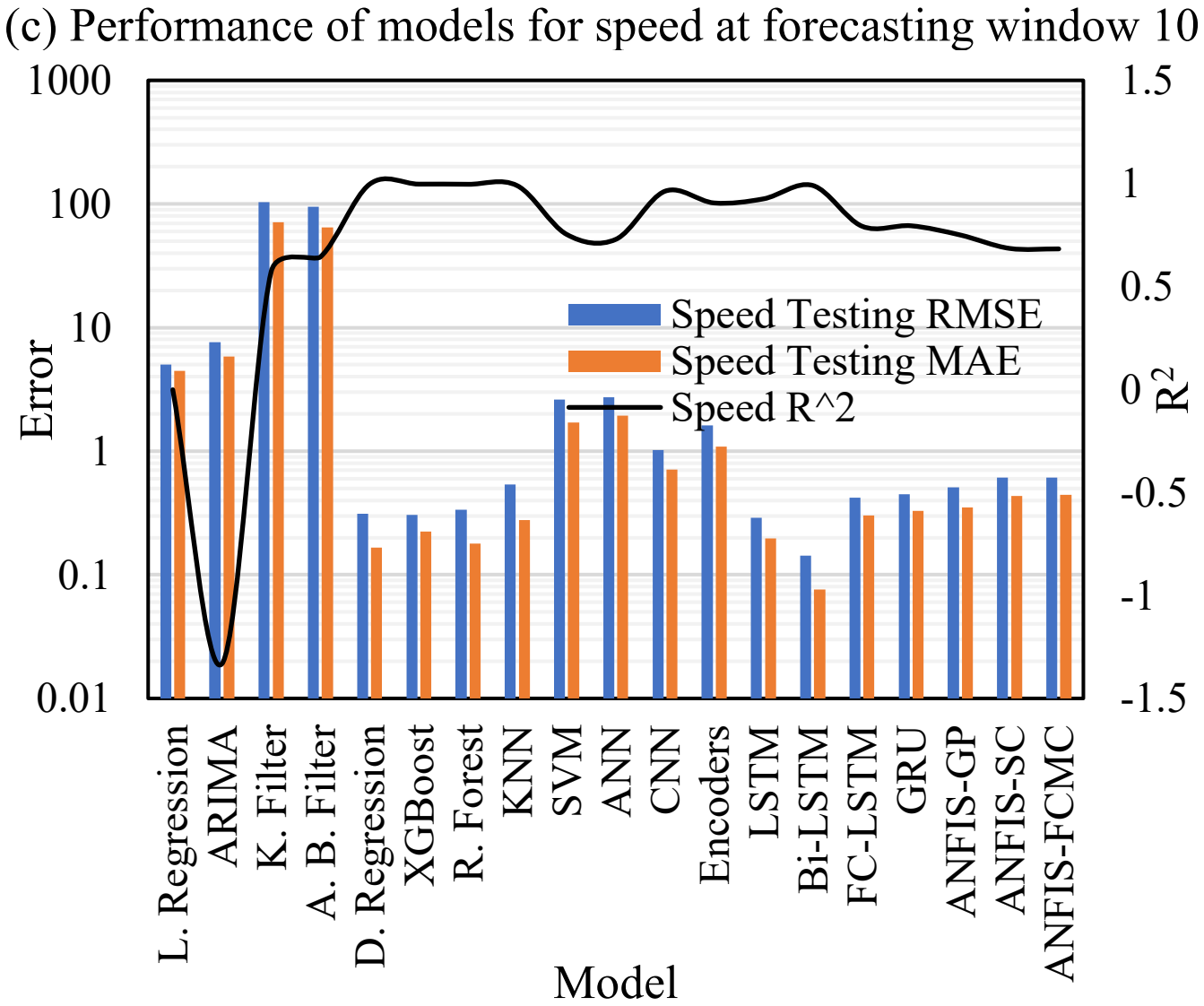

(e) Performance of models for speed at forecasting window 20

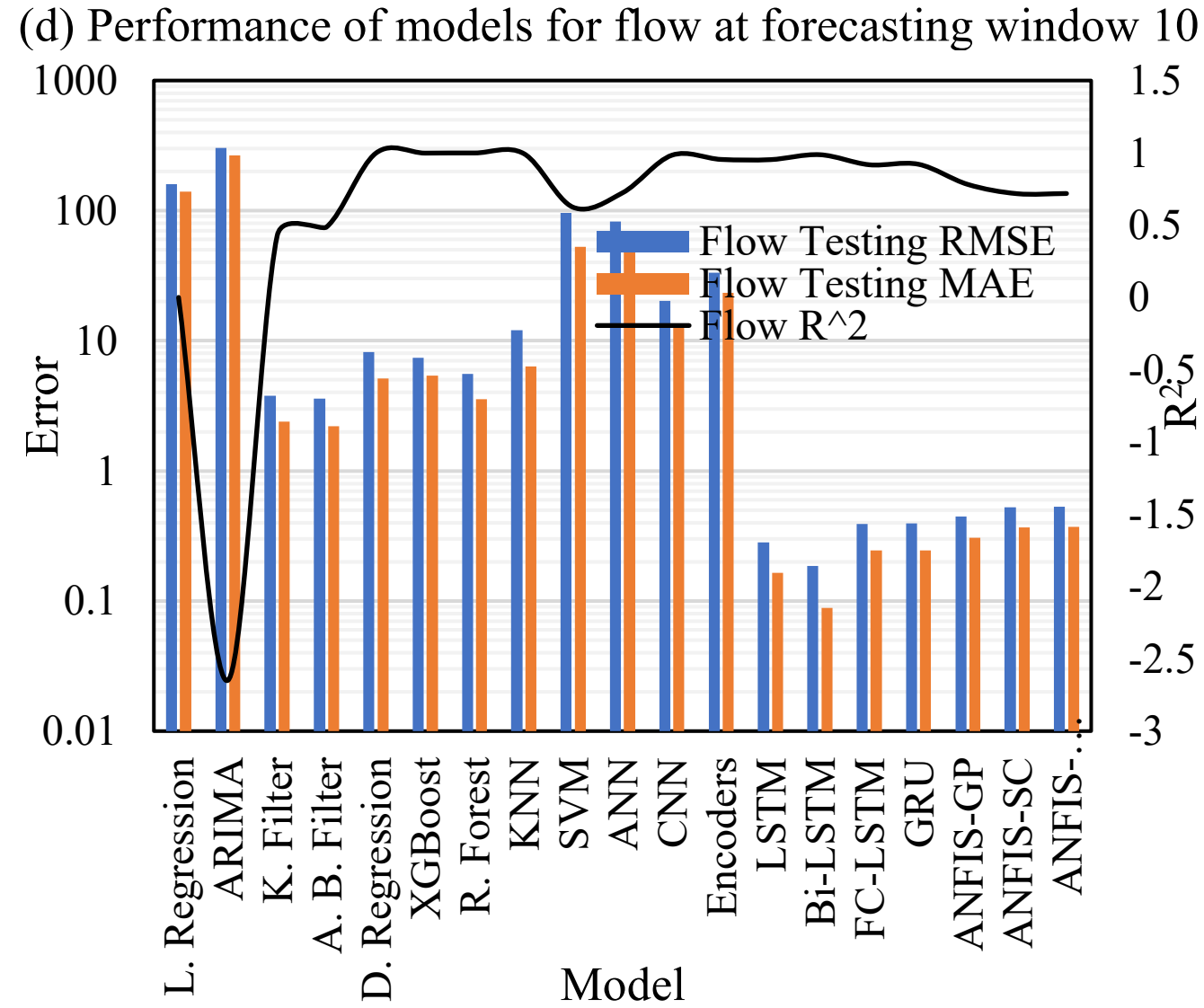

(f) Performance of models for flow at forecasting window 20

**FIGURE 2**. Performance of the various models for forecasting windows 1, 10 and 20

more pronounced, thus increasing the errors of the predicted values. Bi-LSTM, on the other, processes data sequences both in the backward and forward direction thus capturing long range dependencies making it more robust at higher forecasting windows.

Comparing the performance metrics of speed and flow, it can be seen that at any forecasting window, the RMSE and MAE of flow is higher than that of speed. This is mainly because the values of flow are in a higher range between 0-500 vehicles/5 minutes whereas the speed data values are within the speed limit of 70 mph for the road segment. Hence, the errors of the flow models are more pronounced than the speed models. Another reason is that the data extracted from loop detectors gives noisier data for traffic flow than for speed (*21*).

### Model Performance with Increasing Forecasting Windows

The variation of the performance metrics of the machine learning and deep learning models with increasing forecasting windows is shown in **Table 2**. The performance metrics RMSE, MAE and $R^2$ were found to have an exponential trendline as the best fit for each model. Hence, the natural logarithm of the performance metrics were taken and these values are plotted against the increasing forecasting windows as shown in **Table 2**. The line of best fit is now in the form as shown in **Equation 4**.

$$\ln(\text{error}) = \text{m} \,.\, (\text{forecasting window}) + \text{c} \tag{4}$$

The slope (m) is a measure of the robustness of each of the models for traffic speed and flow prediction because it represents how fast the error increases with increasing forecasting window and the intercept (c) is a measure of the initial error. A higher slope indicates a unstable model and a lower slope indicates a robust model. From **Table 2**, it is seen that for speed prediction models, Decision Tree Regression and XGBoost has the highest robustness with a slope of 0.0002 for speed RMSE. However, both of these models have high intercept values indicating high initial error. Among the deep learning models, Bi-LSTM has the smallest slope of 0.0423 indicating its robustness and ANFIS-FCMC has the highest slope of 0.0911 indicating its instability. Comparing the initial errors (intercept) of the deep learning speed models, it is seen that LSTM and Bi-LSTM have relative high intercepts of -2.1846 and -2.7307 and ANFIS-GP has an intercept of -2.109. This is why these models have higher initial errors and therefore, the best performing model at forecasting window 1 for speed was found to be ANFIS-GP in section 3.1. Similar results are also found when comparing the MAE slopes and intercepts of the speed models. The highest MAE slope of the deep learning models is 0.1006 for ANFIS-FCMC which shows its instability and the lowest MAE slope is 0.0359 for Bi-LSTM as also seen with RMSE trends.

Among the ML and DL flow models, the smallest slope is seen to be -0.0014 for Decision Tree Regression RMSE which indicates model robustness. Among the DL models only, an RMSE slope of 0.0454 is seen for Bi-LSTM which indicate its robustness for predicting flow data as discussed in the previous section, where Bi-LSTM was seen to outperform all other models at forecasting window 20. Although Decision Tree Regression has the best robustness, it has a high intercept value of 8.2244 for RMSE and 5.134 for MAE indicating its high initial error. From **Table 2**, it can be seen that the slope of the RMSE and MAE for both speed and flow are positive (except for Decision Tree Regression flow), indicating that the error values increase with increasing forecasting window. The exception for Decision Tree Regression may be because linear model is not fully capturing the trend in data variation.

The slope of the $R^2$ equations for both speed and flow of all models are negative (except Decision Tree Regression flow) because the $R^2$ value decreases with increasing forecasting window. It can be seen that for most ML models, except ANN and SVM, that slope magnitude is less than 0.01 for both speed and flow indicating that at higher forecasting windows, the data variability is less. For most of the DL models, except CNN, encoders, LSTM and Bi-LSTM, the $R^2$ slope is greater than 0.01 indicating poor fit of the data at higher forecasting windows.

**TABLE 2**. Variation of training RMSE, trainign MAE and $R^2$ for increasing forecasting windows

**Decision Tree Regression**

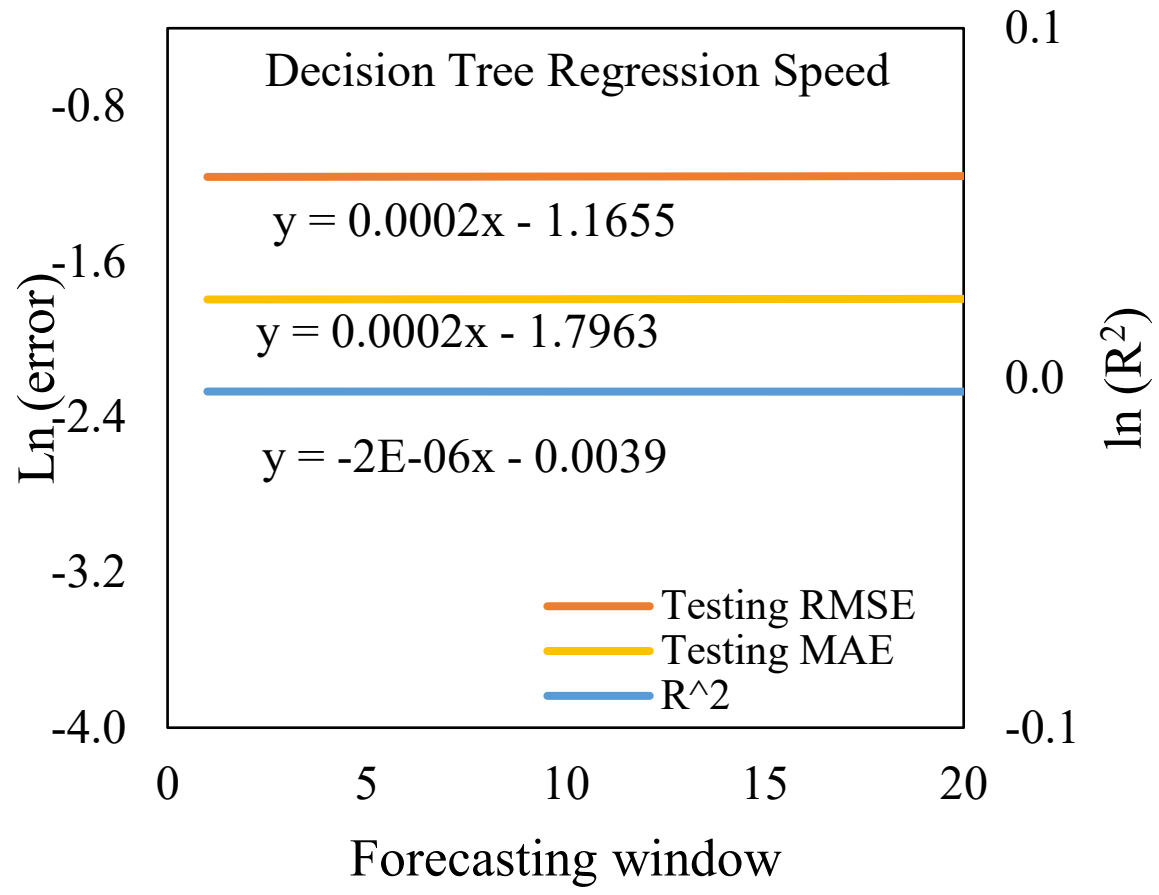

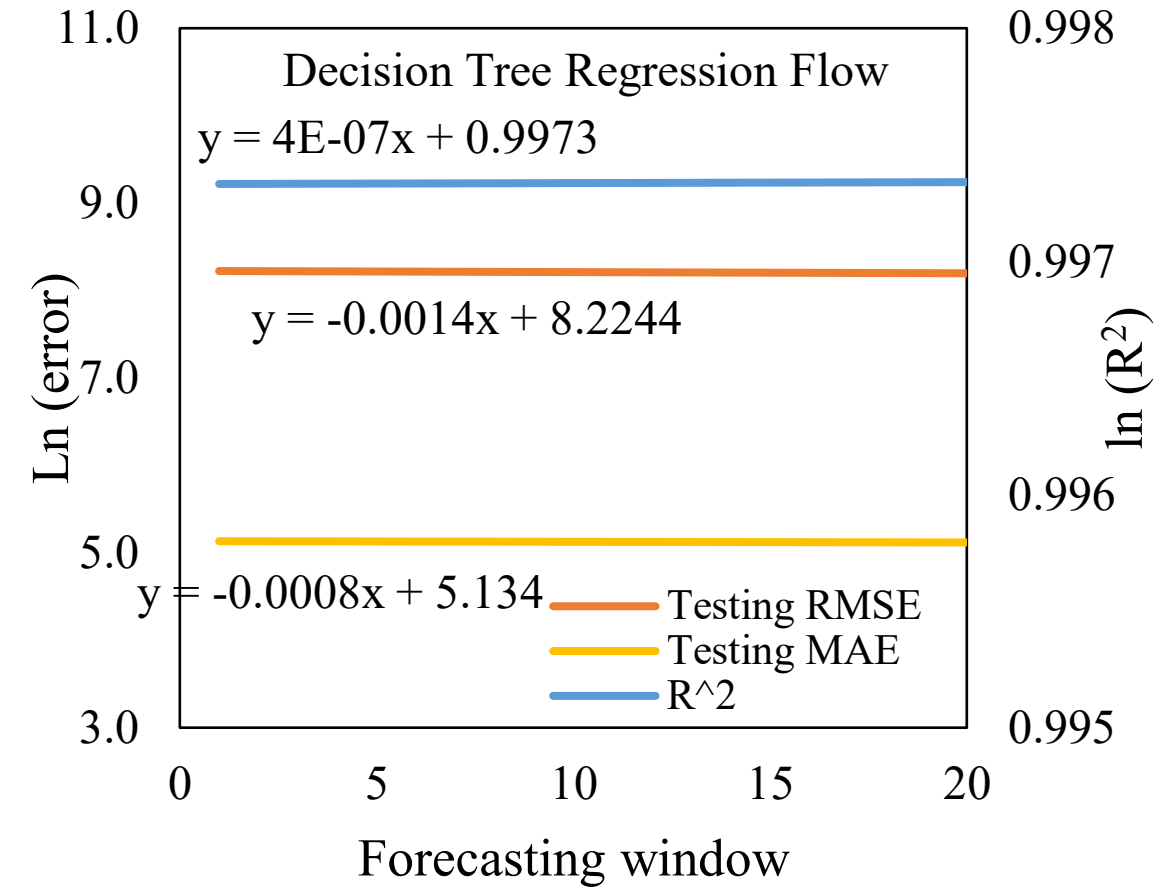

Testing RMSE: $y = 0.0002x - 1.1655$
Testing MAE: $y = 0.0002x - 1.7963$
$R^2$: $y = -0.000002x + 0.0039$

Testing RMSE: $y = -0.0014x + 8.2244$
Testing MAE: $y = -0.0008x + 5.134$
$R^2$: $y = 0.0000004x + 0.9973$

**XGBoost**

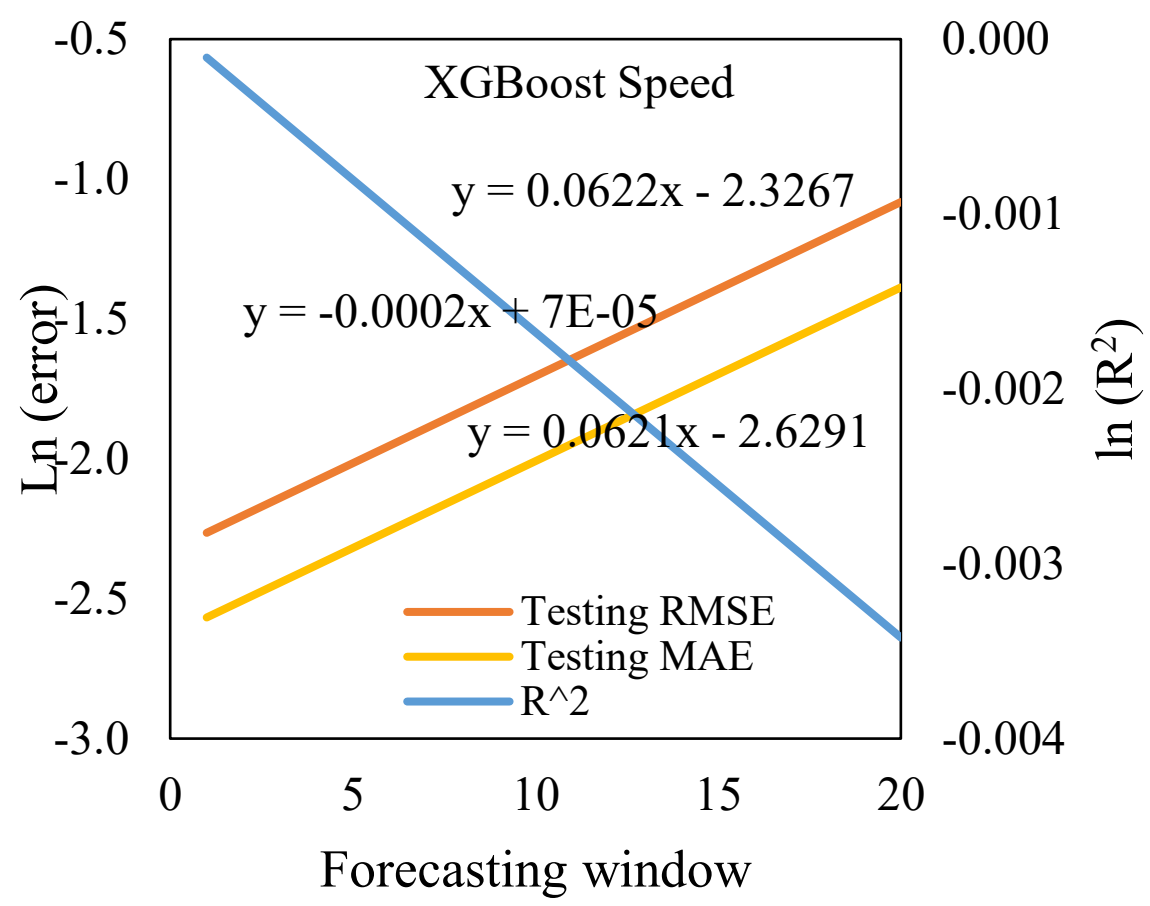

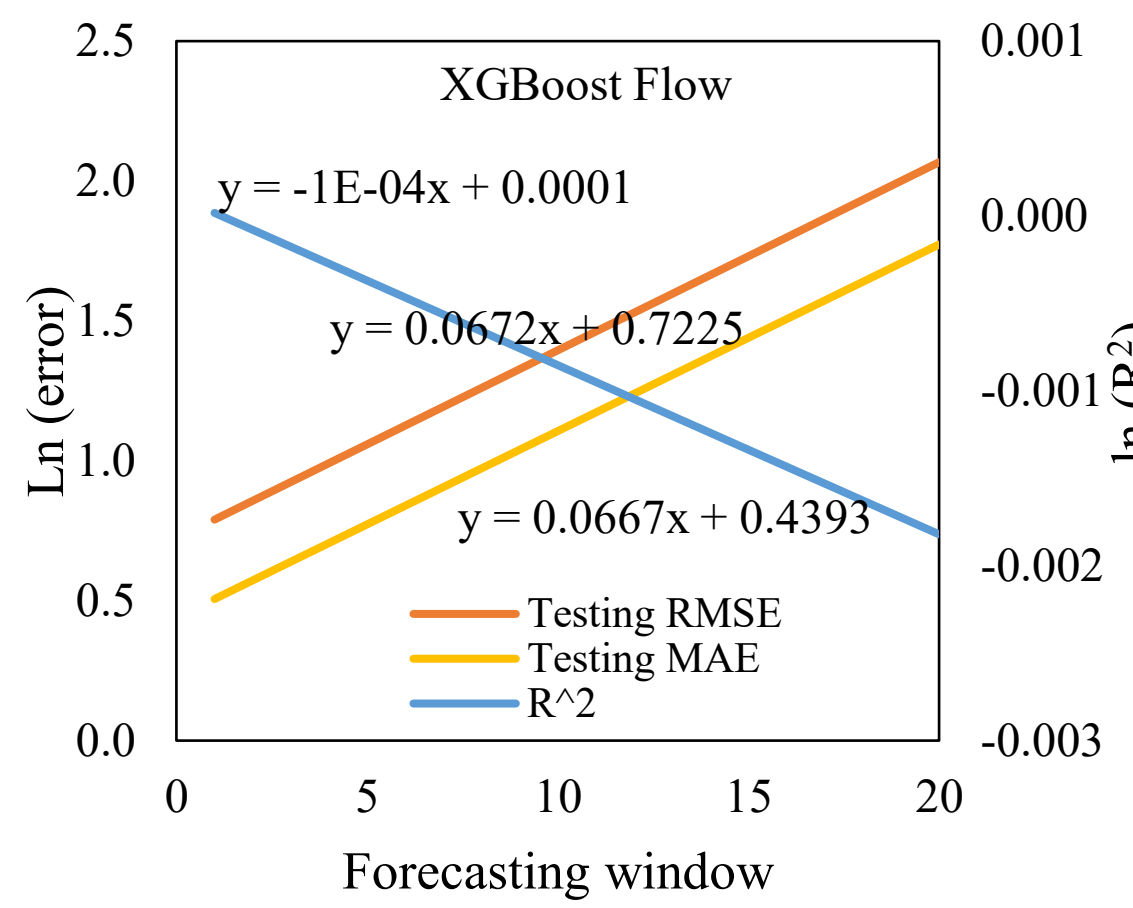

Testing RMSE: $y = 0.0622x - 2.3267$
Testing MAE: $y = 0.0621x - 2.6291$
$R^2$: $y = -0.0002x + 0.00007$

Testing RMSE: $y = 0.0672x + 0.7225$
Testing MAE: $y = 0.0667x + 0.4393$
$R^2$: $y = -0.0001x + 0.0001$

**Random Forest**

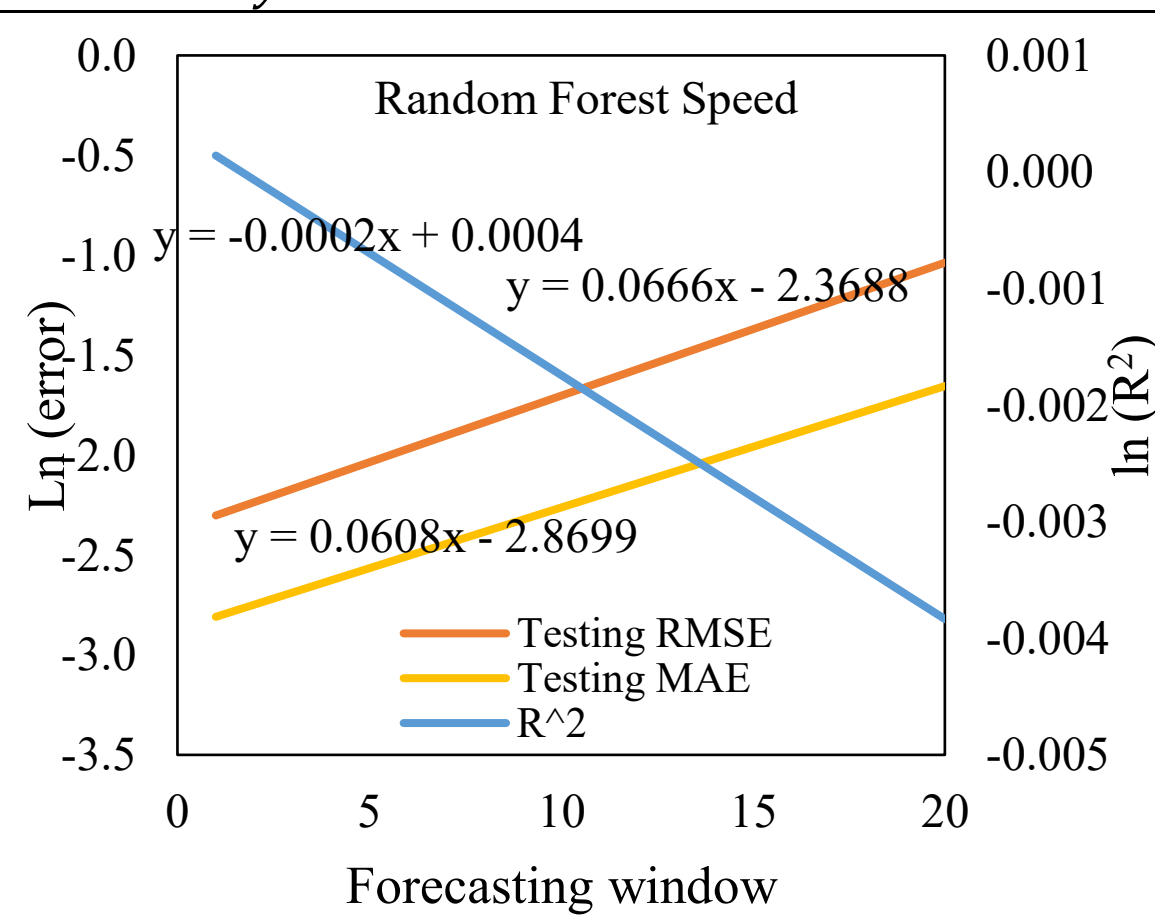

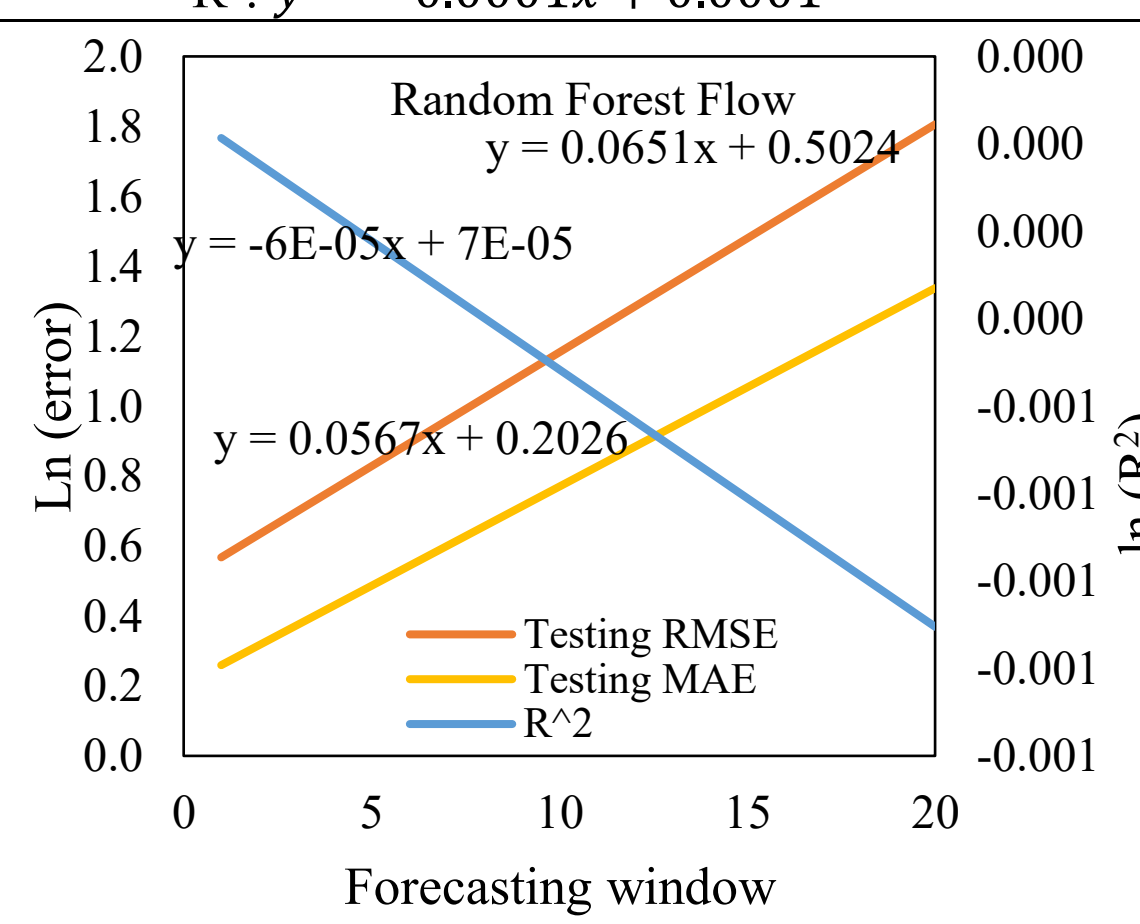

Testing RMSE: $y = 0.0666x - 2.3688$
Testing MAE: $y = 0.0608x - 2.8699$
$R^2$: $y = -0.0002x + 0.0004$

Testing RMSE: $y = 0.0651x + 0.5024$
Testing MAE: $y = 0.0567x + 0.2026$
$R^2$: $y = -0.00006x + 0.00007$

**KNN**

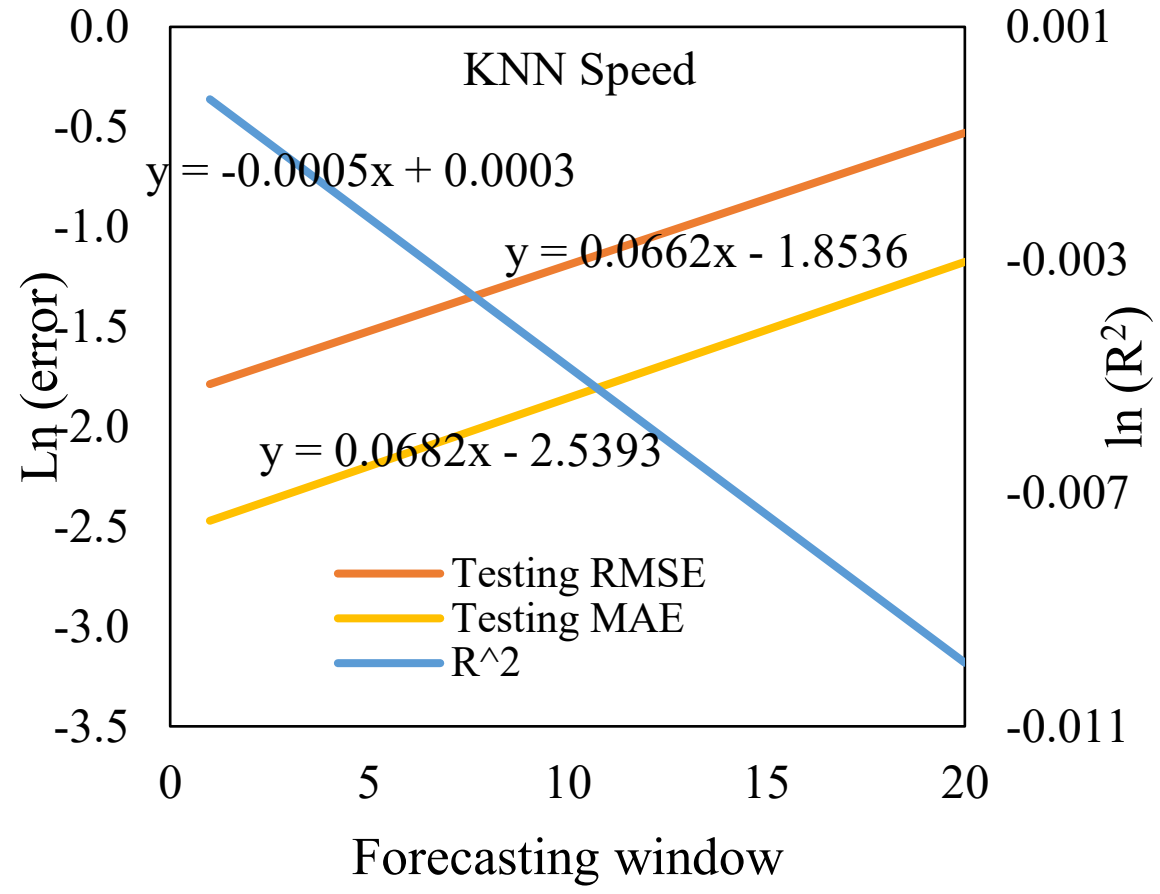

Testing RMSE: $y = 0.0662x - 1.8536$
Testing MAE: $y = 0.0682x - 2.5393$
$R^2$: $y = -0.0005x + 0.0003$

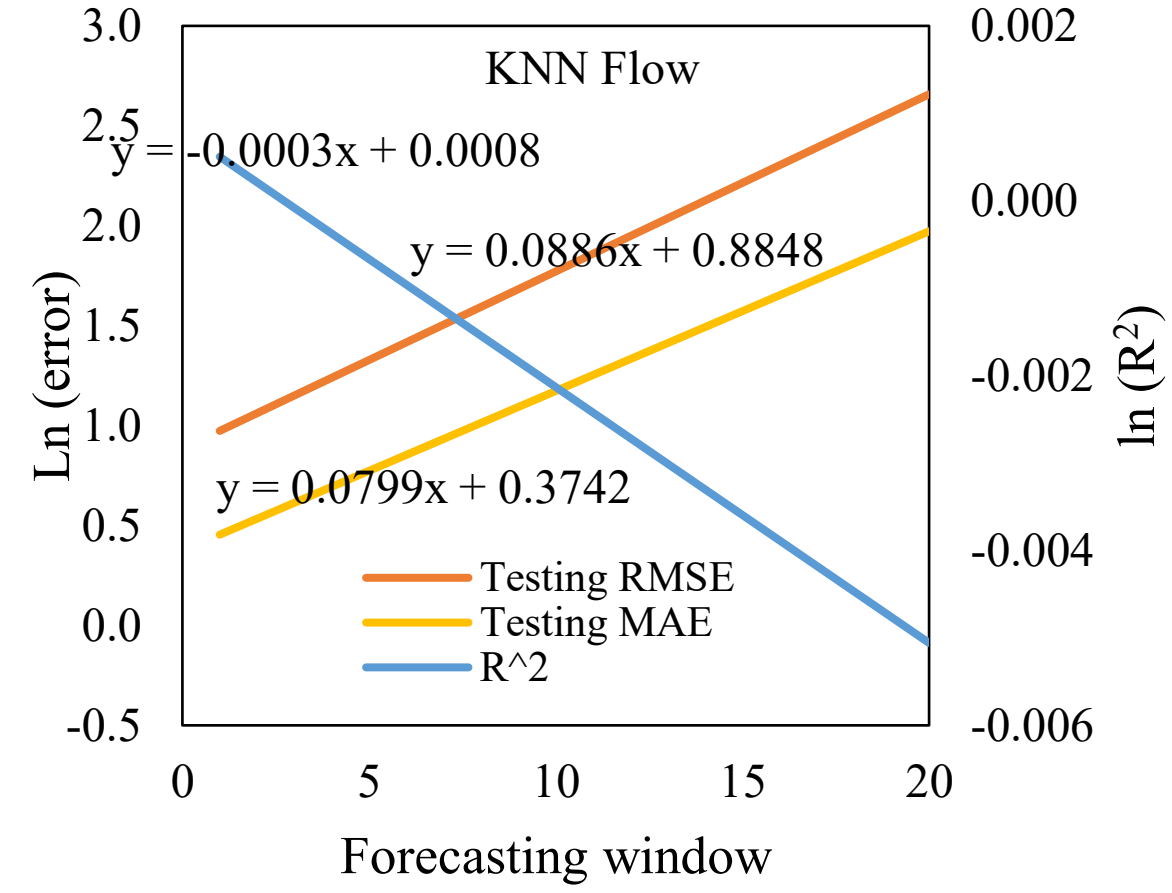

Testing RMSE: $y = 0.0886x + 0.8848$
Testing MAE: $y = 0.0799x + 0.3742$
$R^2$: $y = -0.0003x + 0.0008$

**SVM**

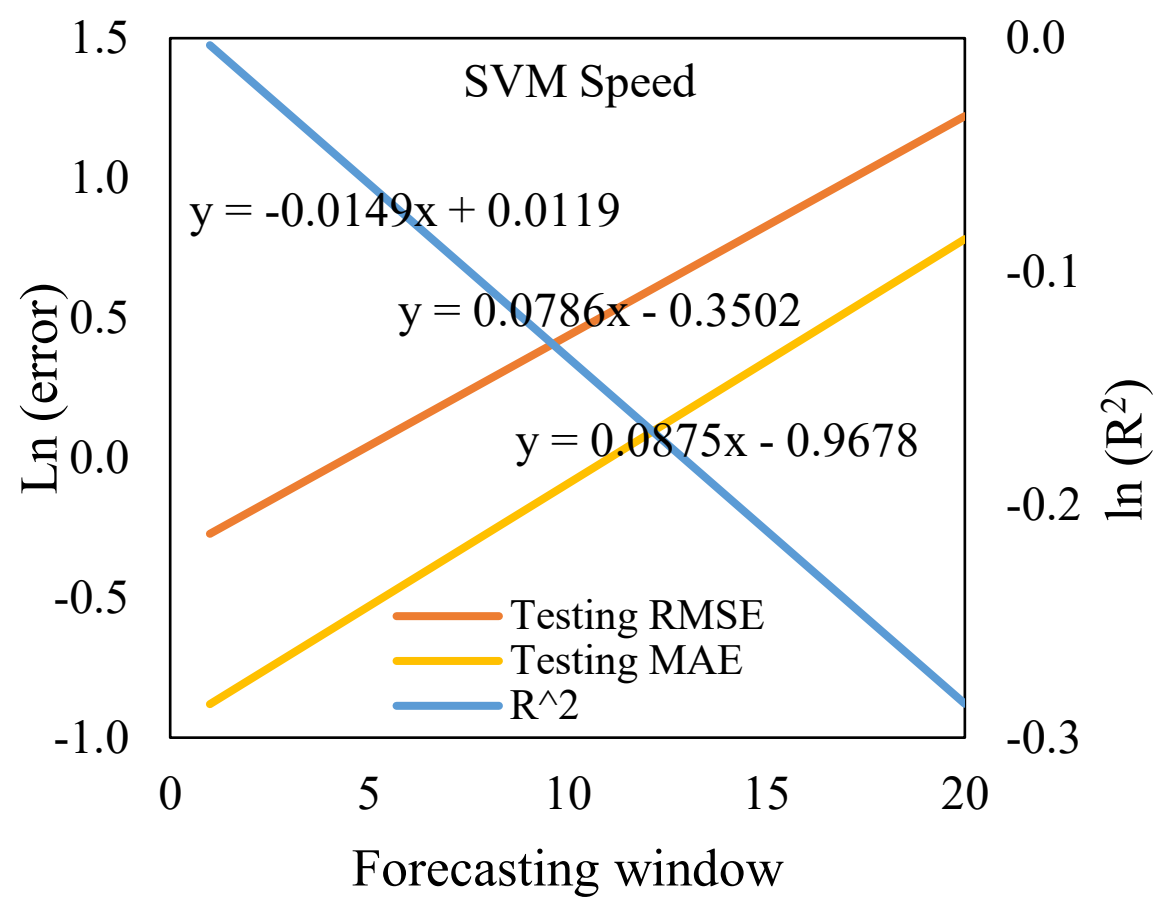

Testing RMSE: $y = 0.0786x - 0.3502$
Testing MAE: $y = 0.0875x - 0.9678$
$R^2$: $y = -0.0149x + 0.0119$

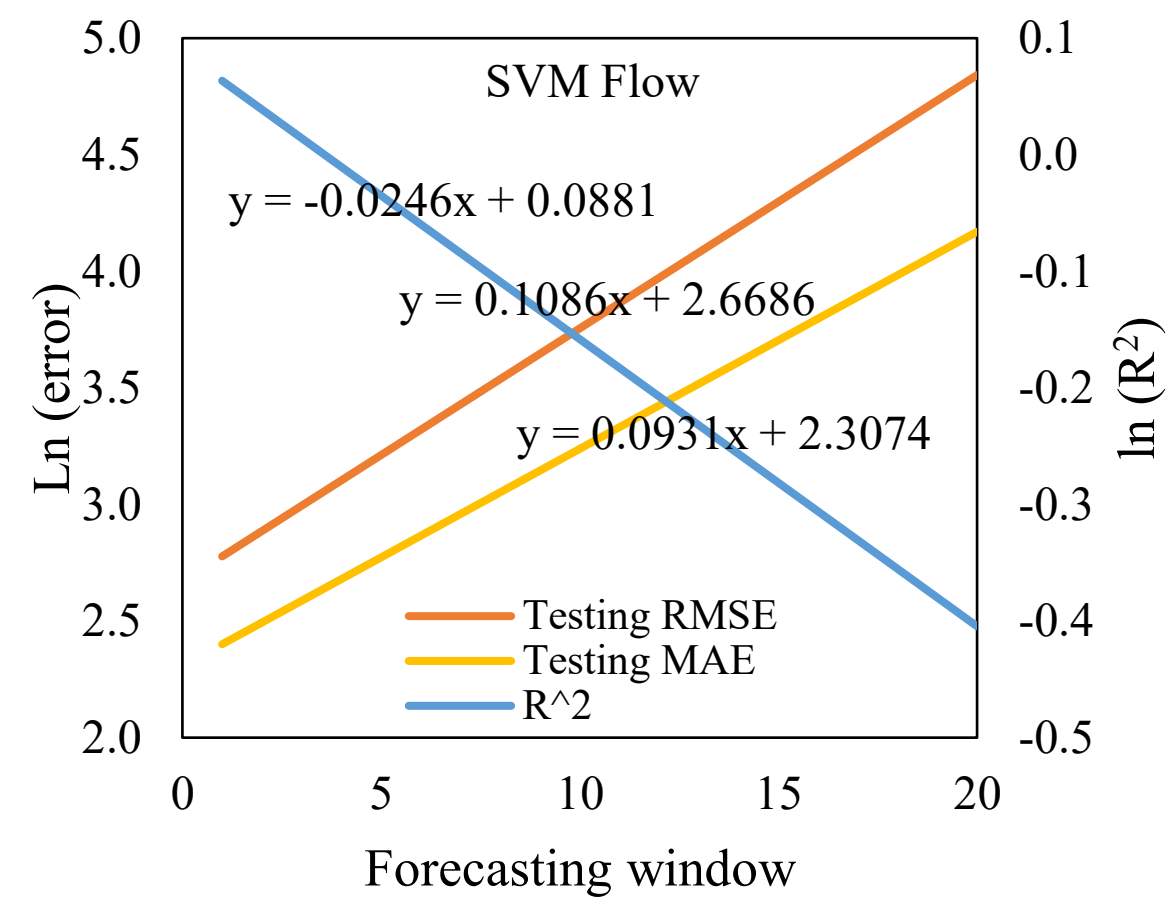

Testing RMSE: $y = 0.1086x + 2.6686$
Testing MAE: $y = 0.0931x + 2.3074$
$R^2$: $y = -0.0246x + 0.0881$

**ANN**

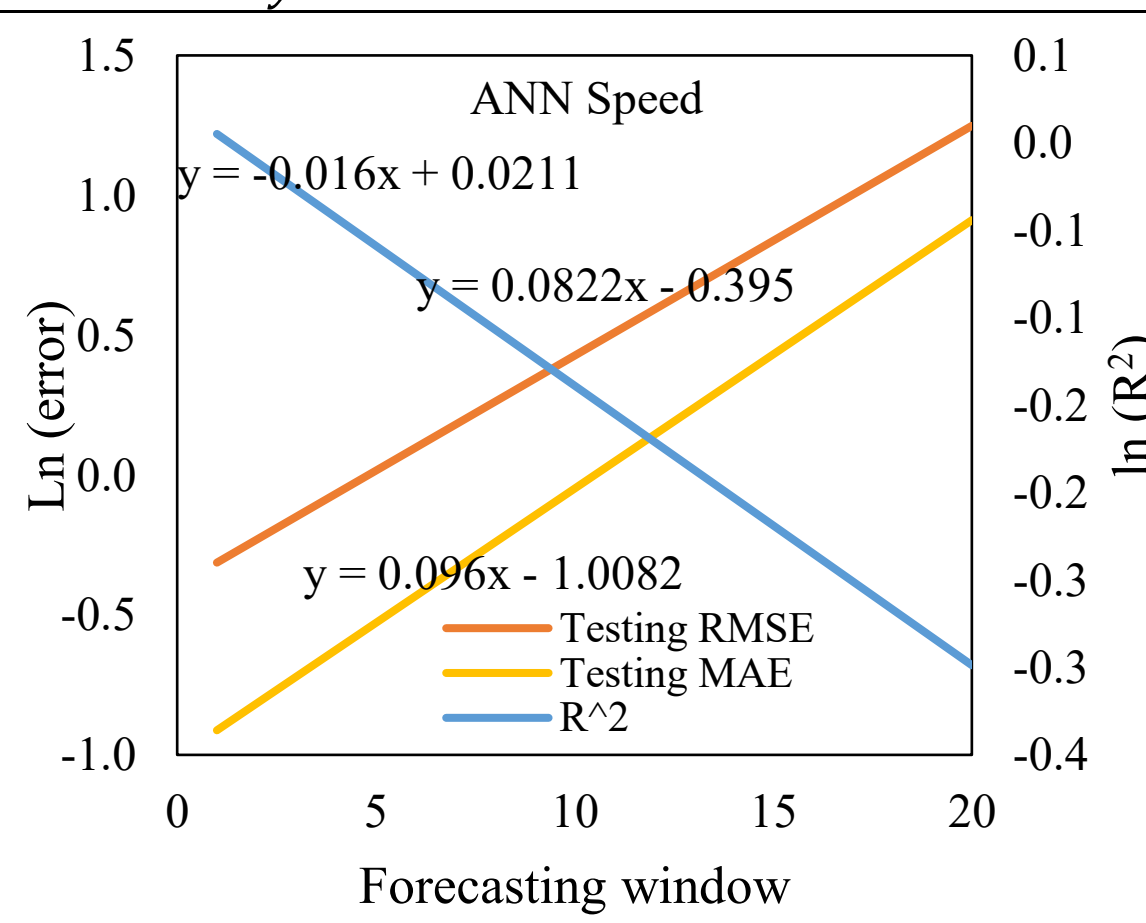

Testing RMSE: $y = 0.0822x - 0.395$
Testing MAE: $y = 0.096x - 1.0082$
$R^2$: $y = -0.016x + 0.0211$

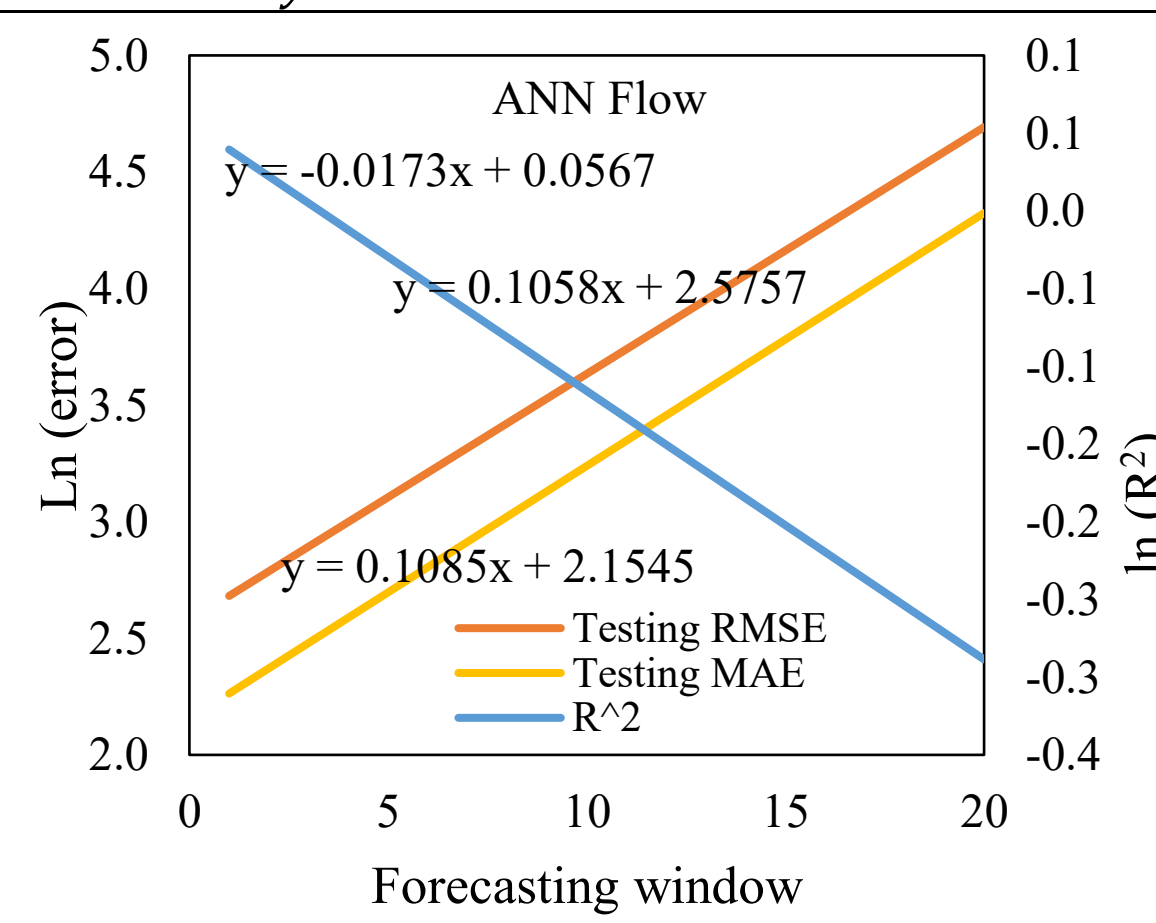

Testing RMSE: $y = 0.1058x + 2.5757$
Testing MAE: $y = 0.1085x + 2.1545$
$R^2$: $y = -0.0173x + 0.0567$

CNN

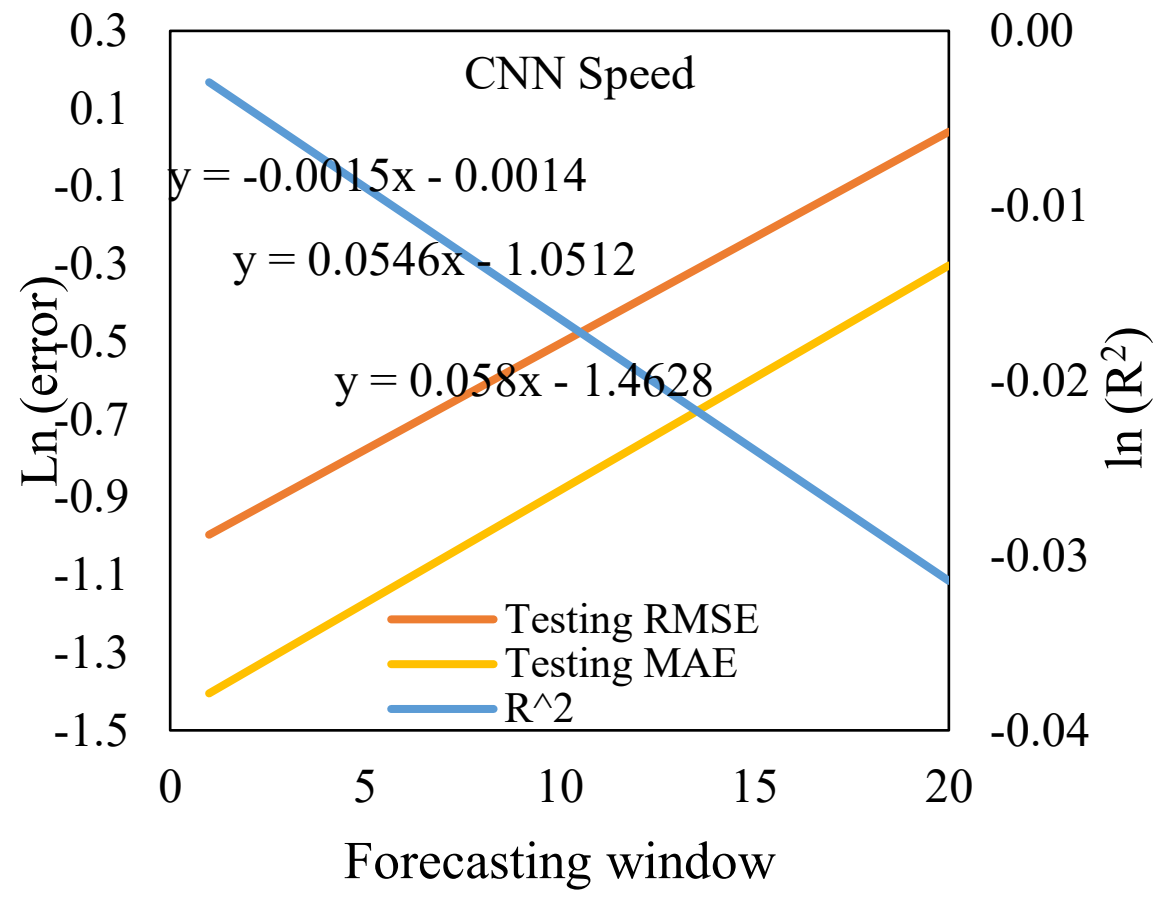

Testing RMSE: $y = 0.0546x - 1.0512$
Testing MAE: $y = 0.058x - 1.4628$
$R^2$: $y = -0.0015x + 0.0014$

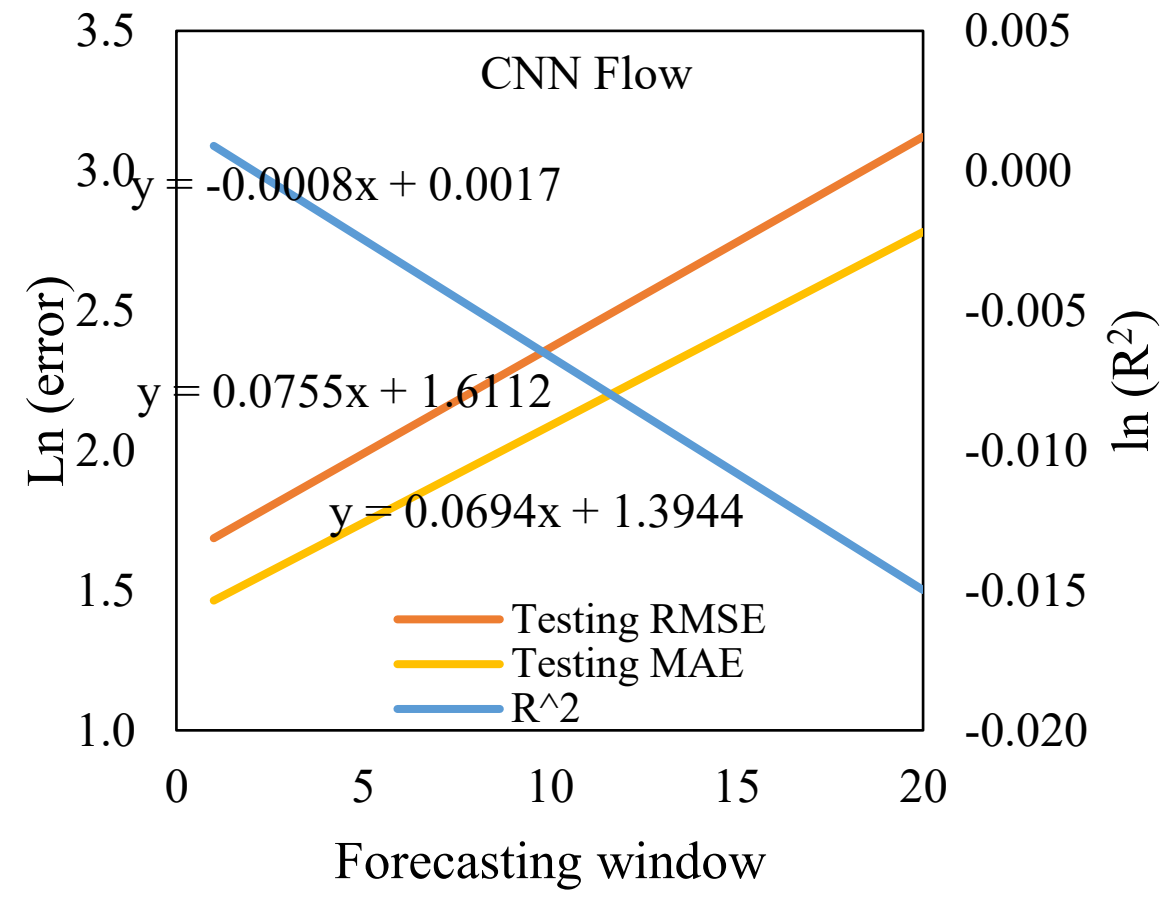

Testing RMSE: $y = 0.0755x + 1.6112$
Testing MAE: $y = 0.0694x + 1.3944$
$R^2$: $y = -0.0008x + 0.0017$

Encoders

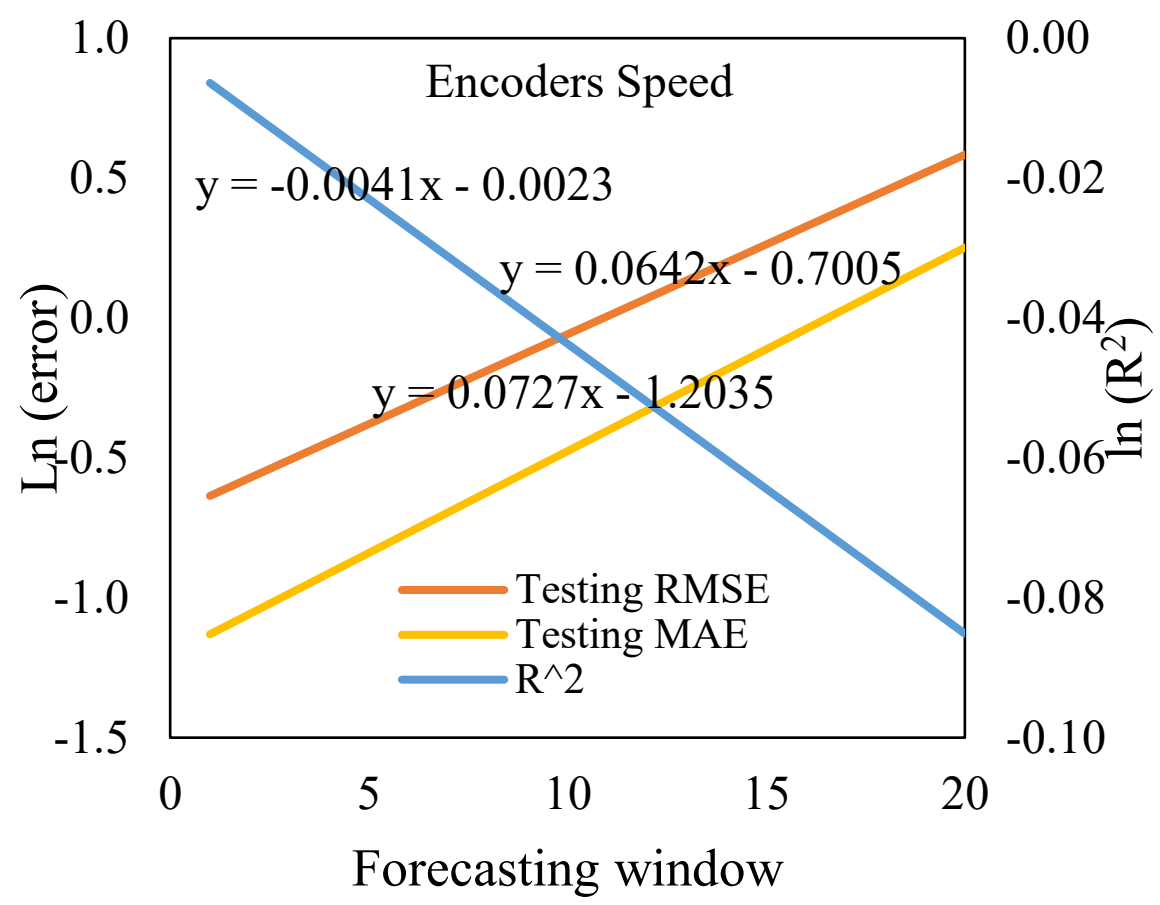

Testing RMSE: $y = 0.0546x - 1.0512$
Testing MAE: $y = 0.058x - 1.4628$
$R^2$: $y = -0.0015x + 0.0014$

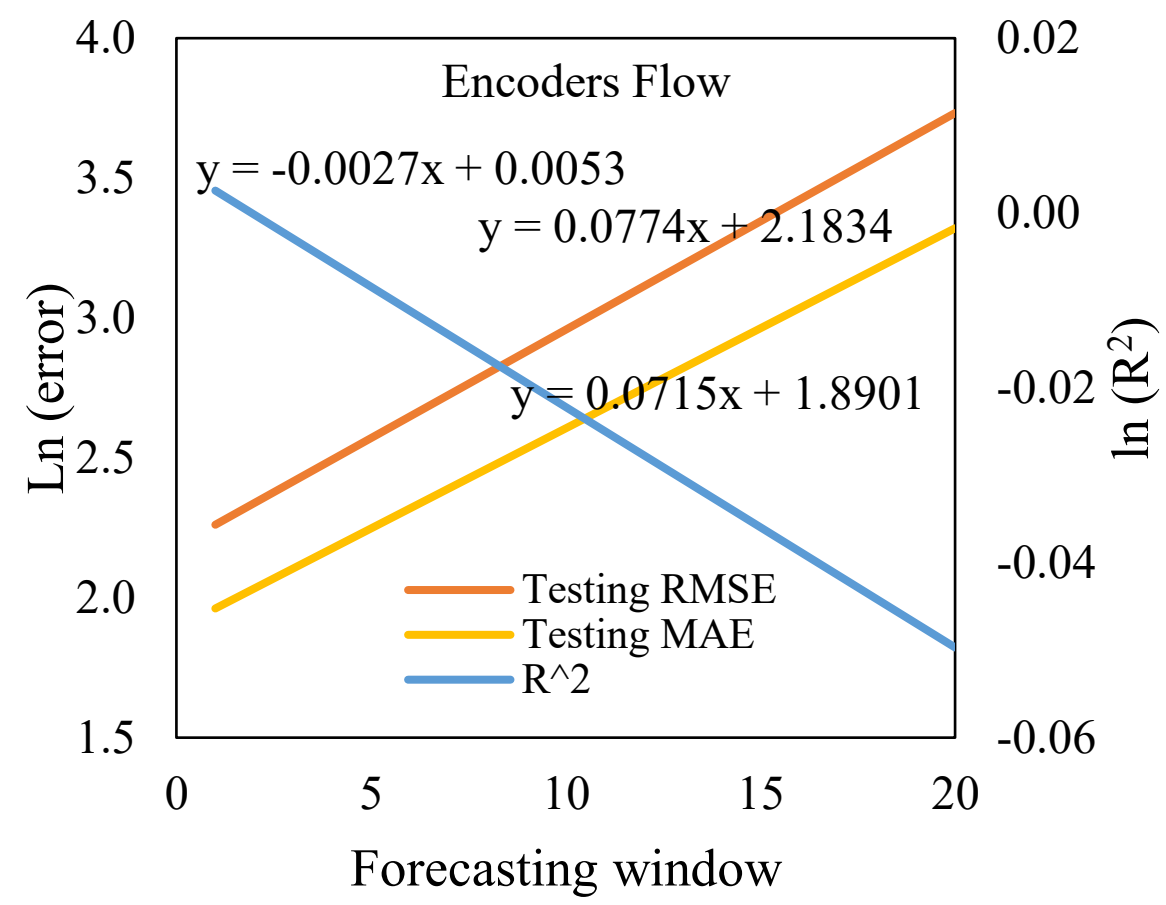

Testing RMSE: $y = 0.0774x + 2.1834$
Testing MAE: $y = 0.0715x + 1.8901$
$R^2$: $y = -0.0027x + 0.0053$

LSTM

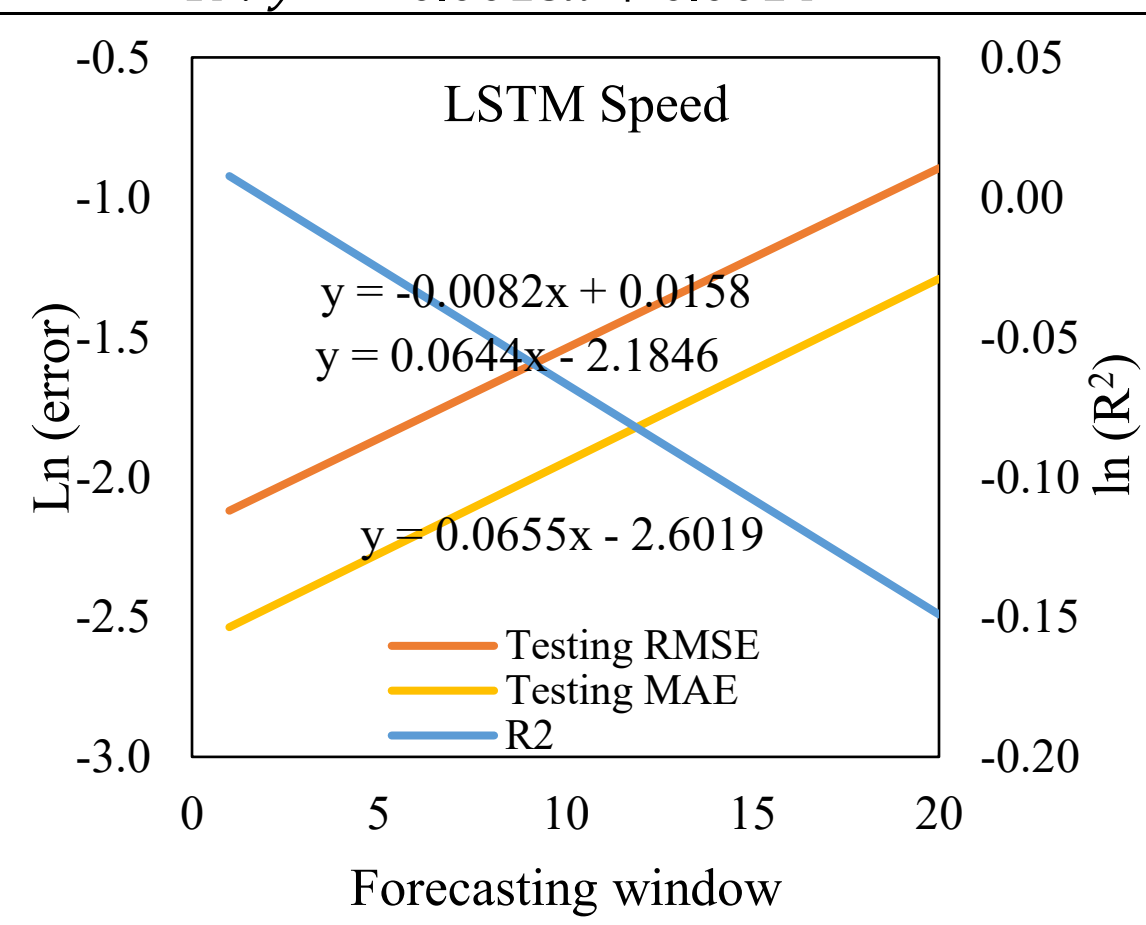

Testing RMSE: $y = 0.0644x - 2.1846$
Testing MAE: $y = 0.0655x - 2.6019$
$R^2$: $y = -0.0082x + 0.0158$

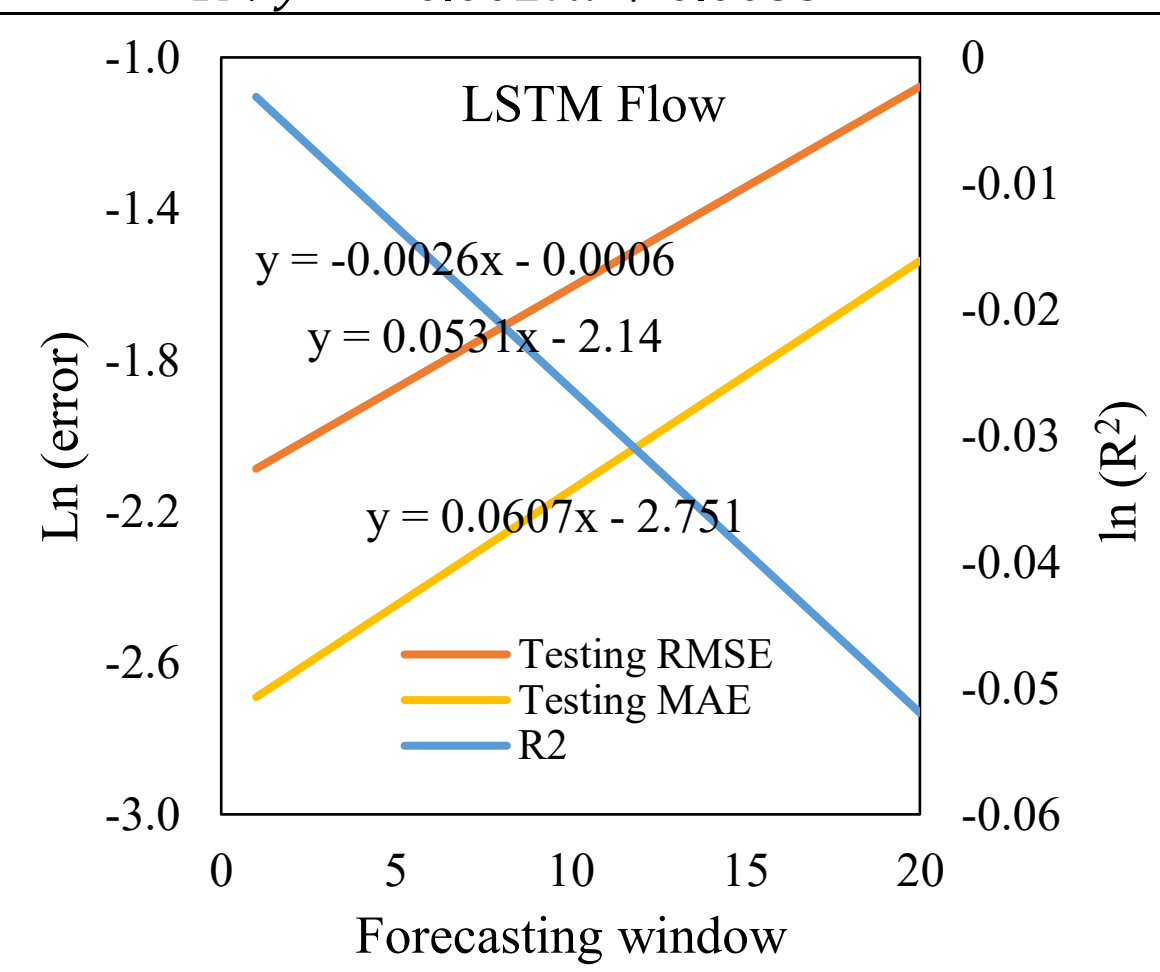

Testing RMSE: $y = 0.0607x - 2.751$
Testing MAE: $y = 0.0531x - 2.14$
$R^2$: $y = -0.0026x - 0.0006$

| | | |
|---|---|---|
| Bi-LSTM | 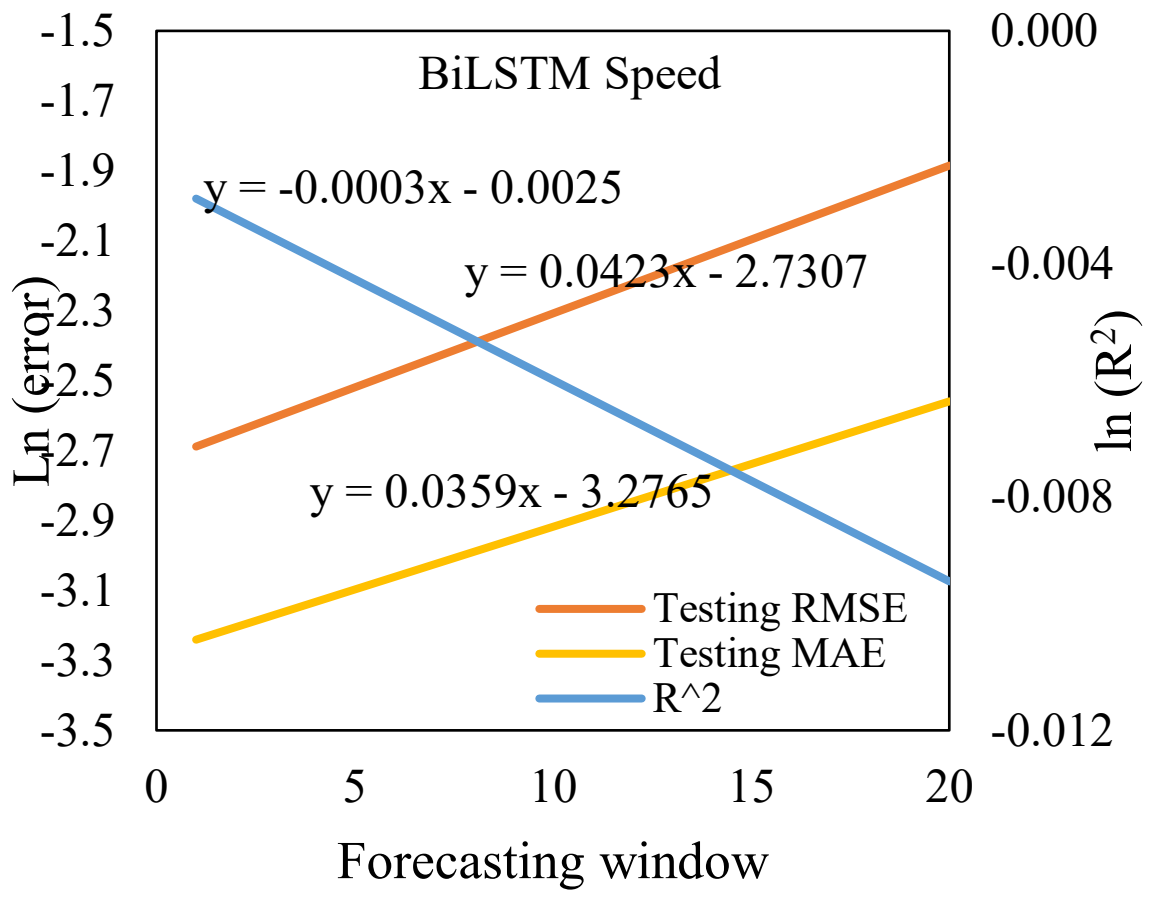<br><br>Testing RMSE: $y = 0.0423x - 2.7307$<br>Testing MAE: $y = 0.0359x - 3.2765$<br>$R^2$: $y = -0.0003x - 0.0025$ | 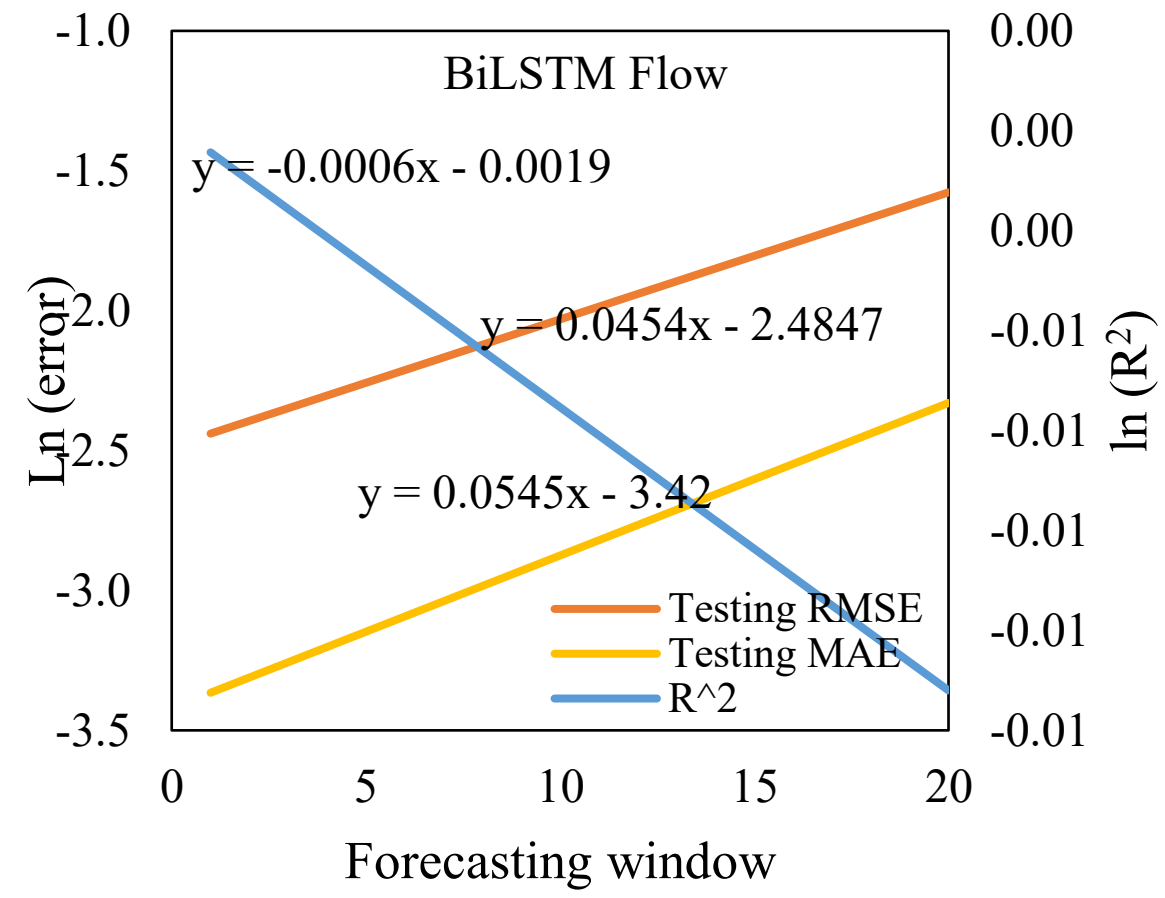<br><br>Testing RMSE: $y = 0.0454x - 2.4847$<br>Testing MAE: $y = 0.05454x - 3.42$<br>$R^2$: $y = -0.0006x - 0.0019$ |
| FC-LSTM | 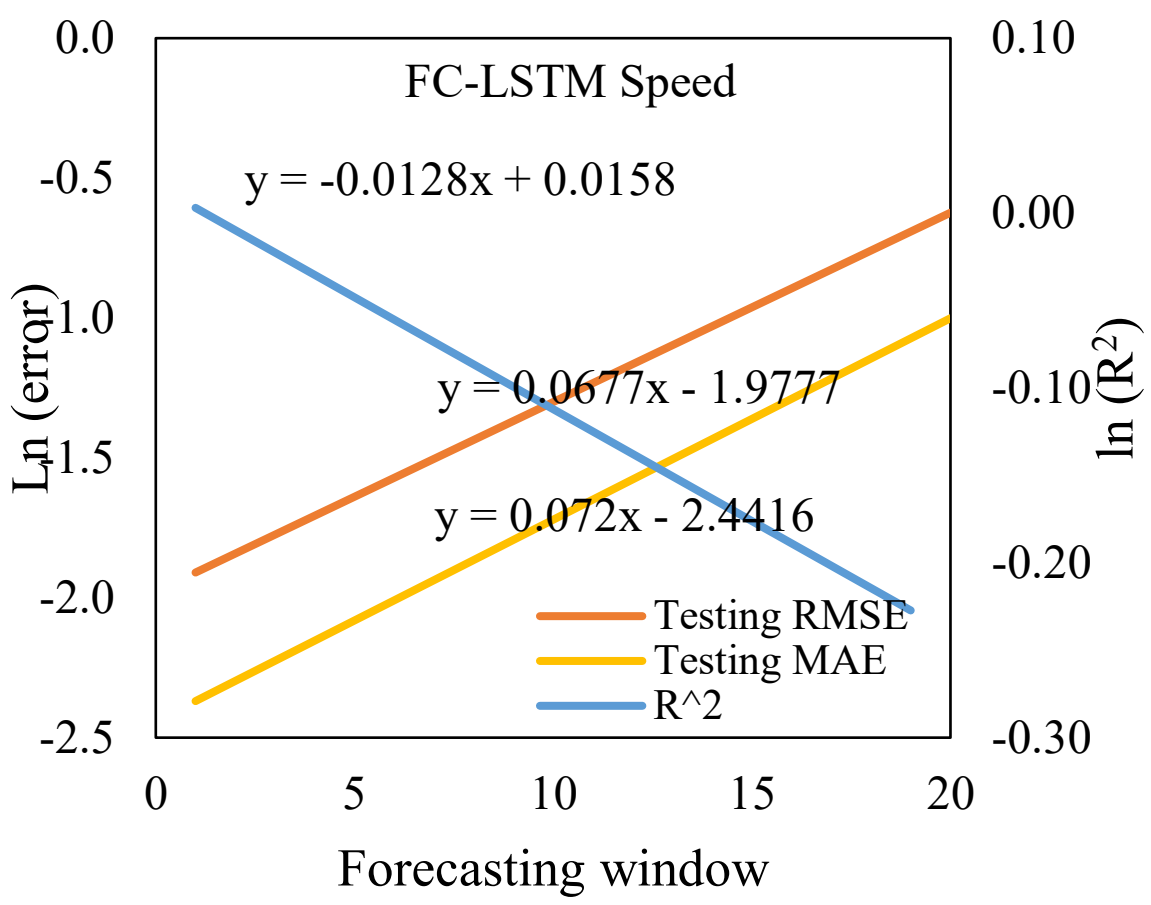<br><br>Testing RMSE: $y = 0.0677x - 1.9777$<br>Testing MAE: $y = 0.072x - 2.4416$<br>$R^2$: $y = -0.0128x - 0.0158$ | 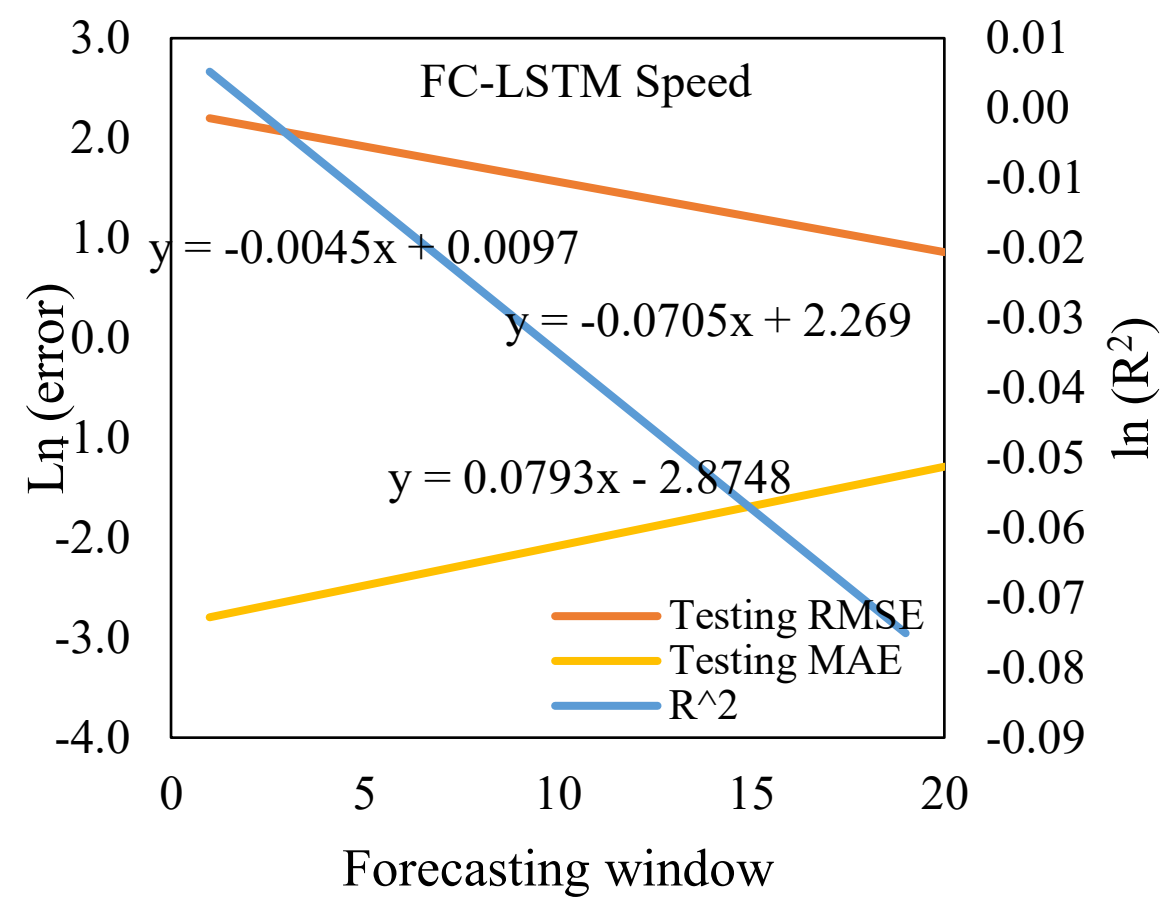<br><br>Testing RMSE: $y = -0.0705x + 2.269$<br>Testing MAE: $y = 0.0793x - 2.8748$<br>$R^2$: $y = -0.0045x + 0.0097$ |
| GRU | 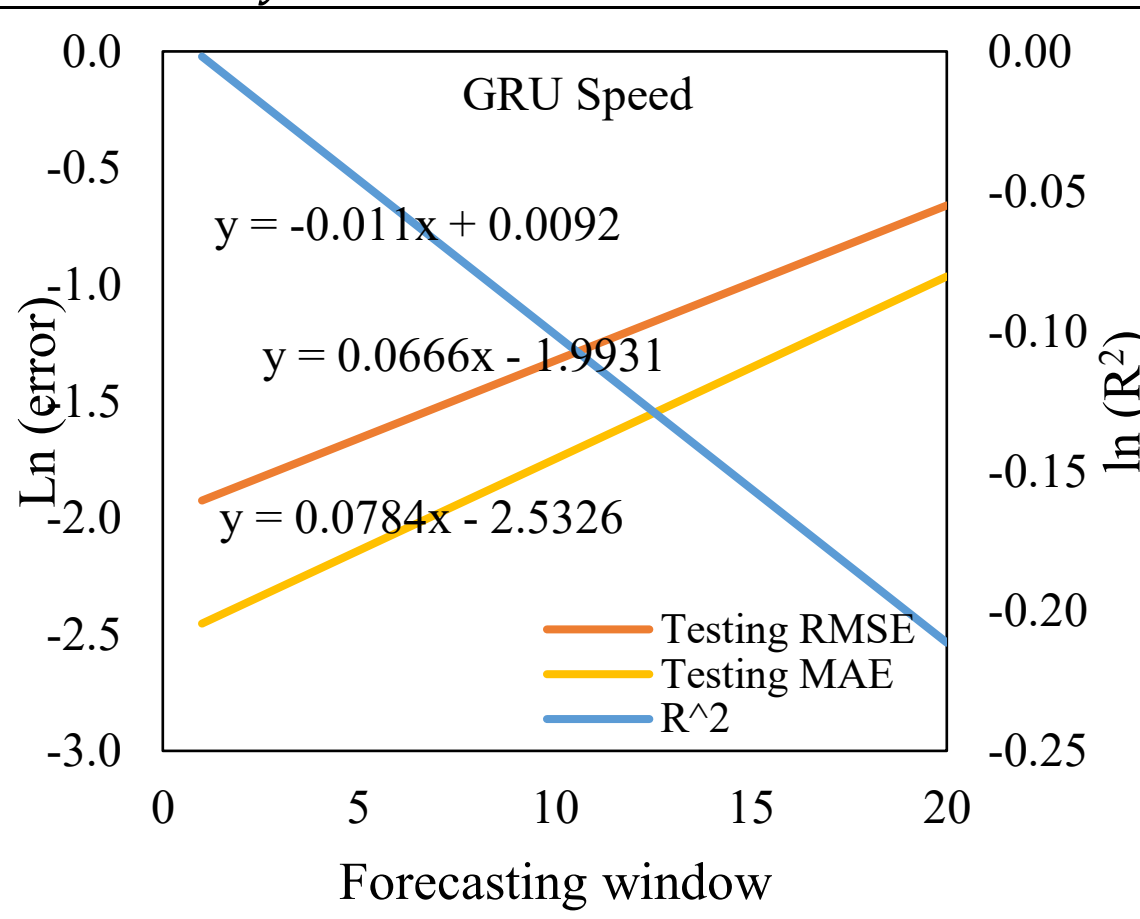<br><br>Testing RMSE: $y = 0.0666x - 1.9931$<br>Testing MAE: $y = 0.0784x - 2.5326$<br>$R^2$: $y = -0.011x + 0.0092$ | 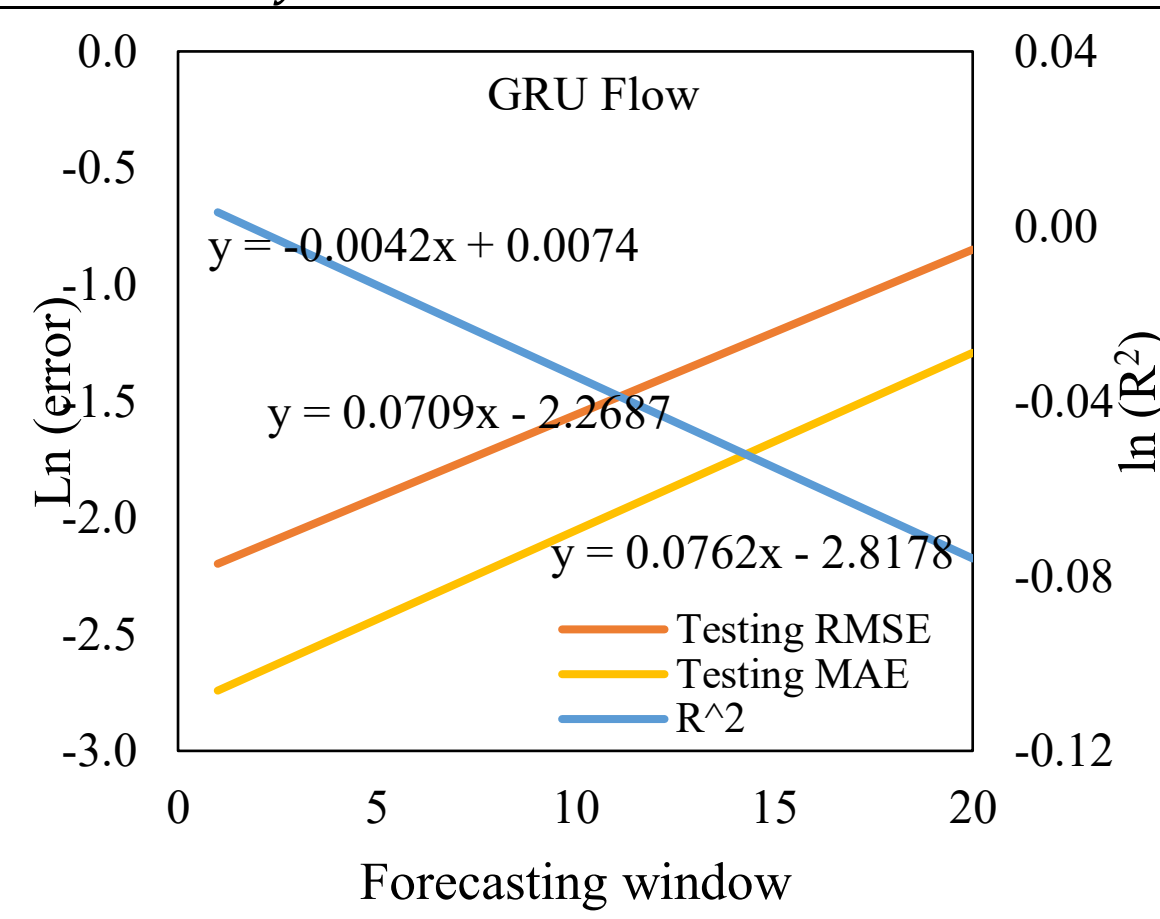<br><br>Testing RMSE: $y = 0.0709x - 2.2687$<br>Testing MAE: $y = 0.0762x - 2.8178$<br>$R^2$: $y = -0.0042x + 0.0074$ |

**ANFIS-GP**

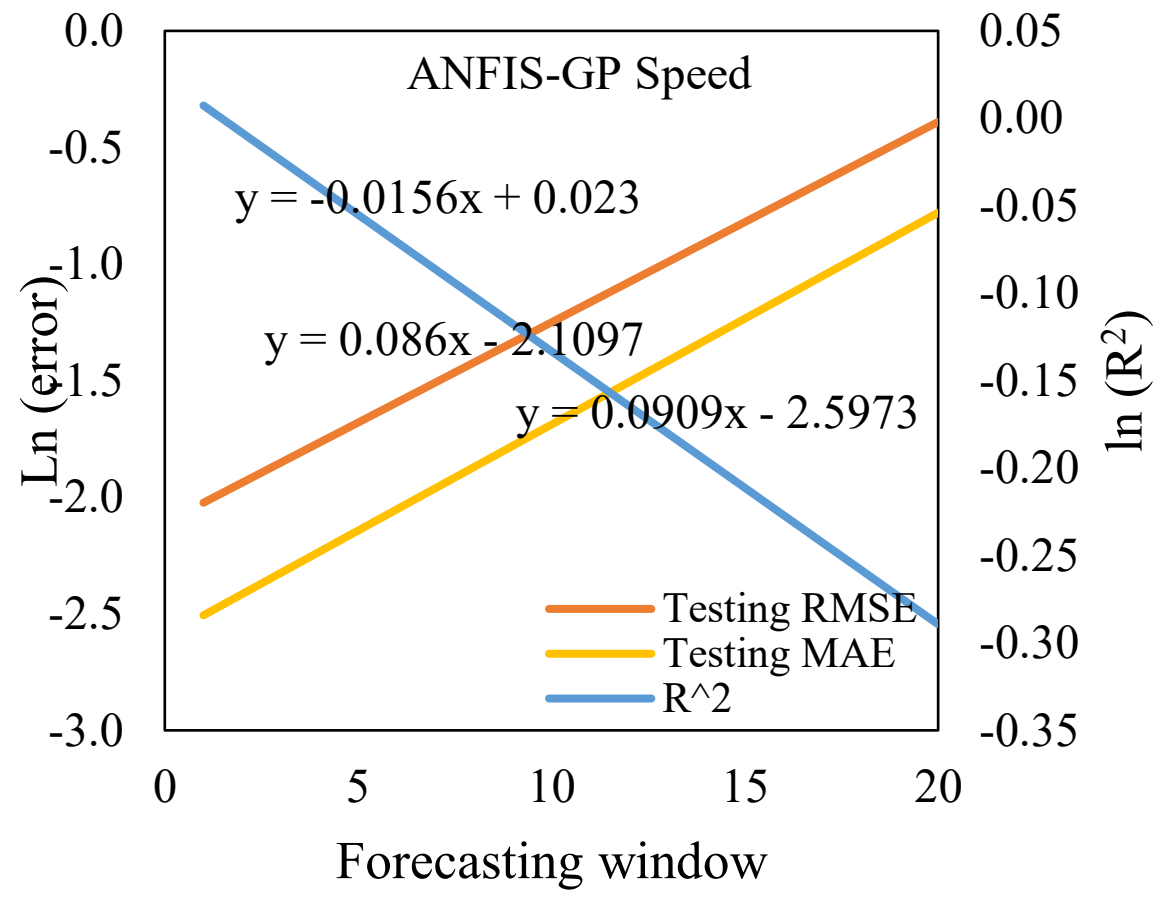

Testing RMSE: $y = 0.086x - 2.1097$
Testing MAE: $y = 0.0909x - 2.5973$
$R^2$: $y = -0.0156x + 0.023$

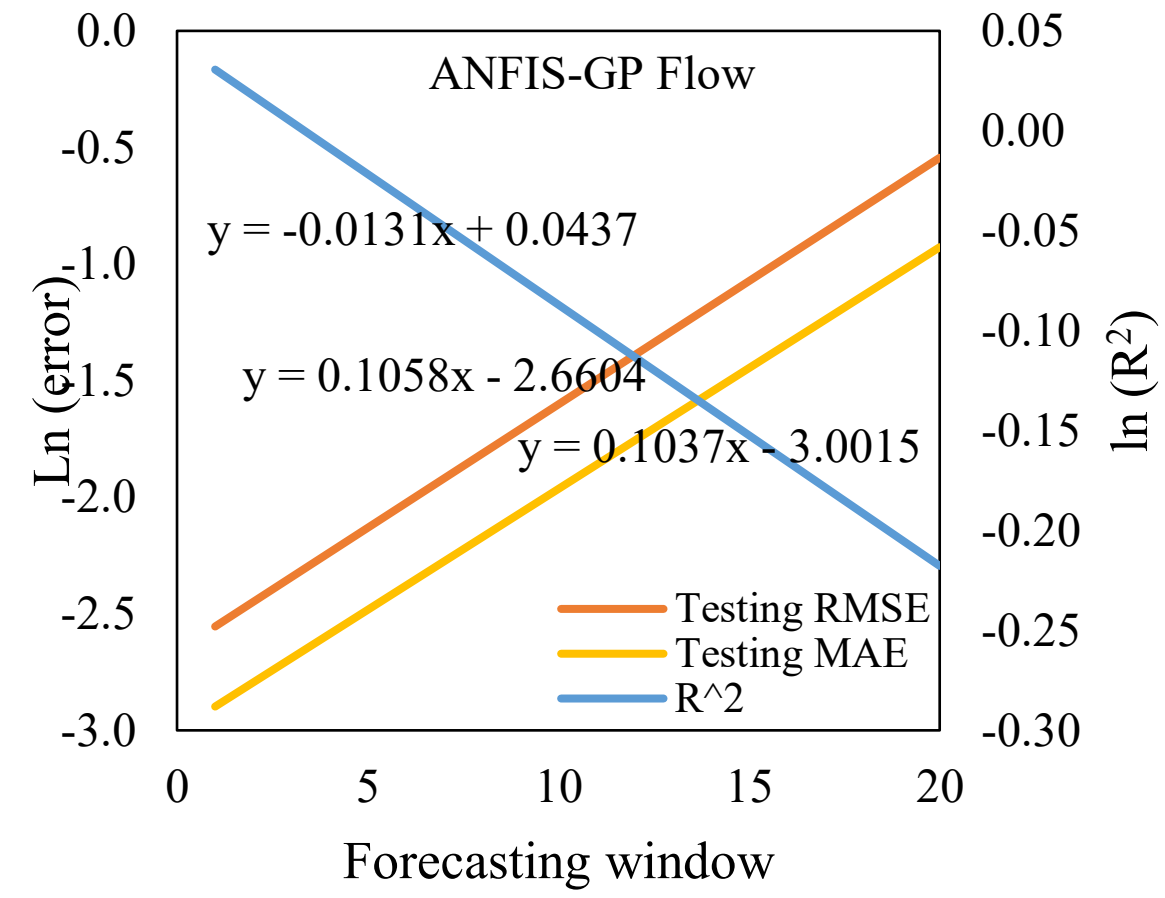

Testing RMSE: $y = 0.1058x - 2.6604$
Testing MAE: $y = 0.1037x - 3.0015$
$R^2$: $y = -0.0131x + 0.0437$

**ANFIS-SC**

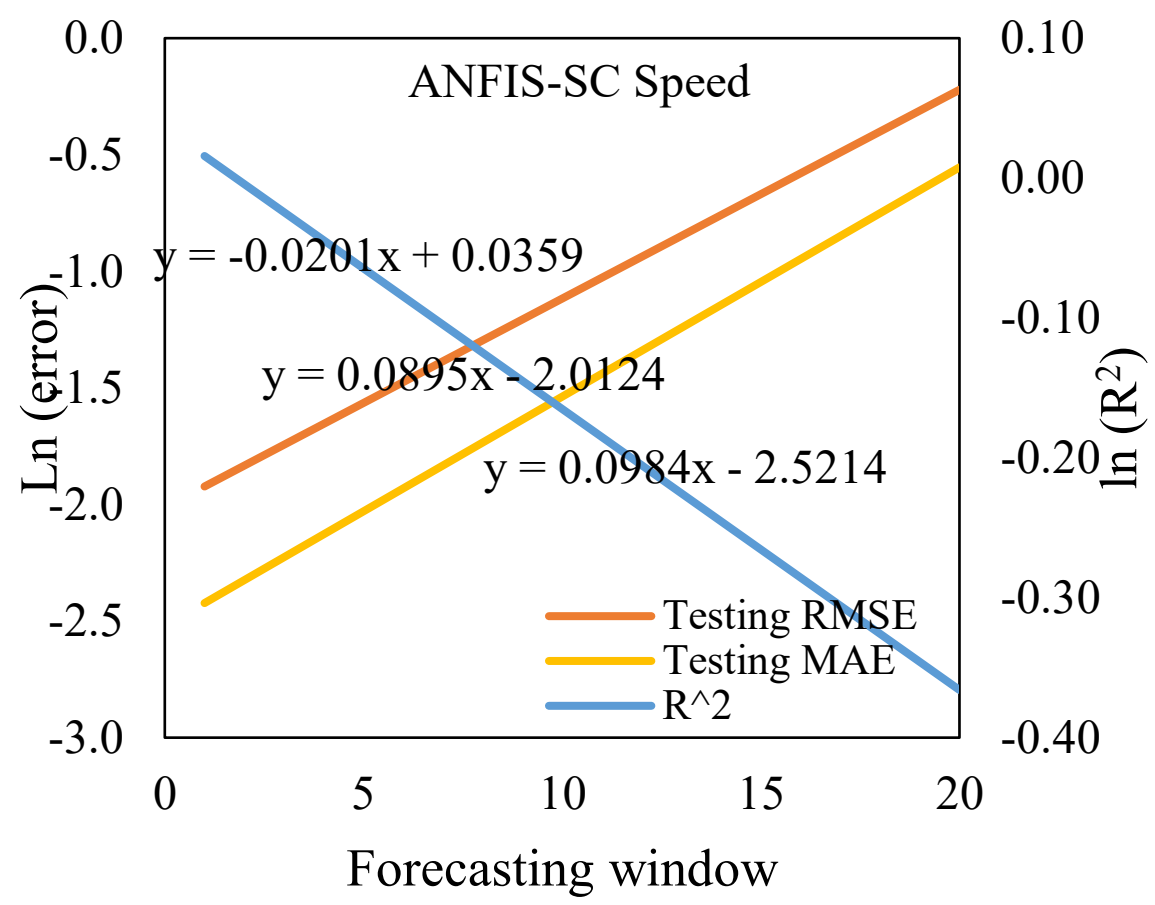

Testing RMSE: $y = 0.0895x - 2.0124$
Testing MAE: $y = 0.0984x - 2.5214$
$R^2$: $y = -0.0201x + 0.0359$

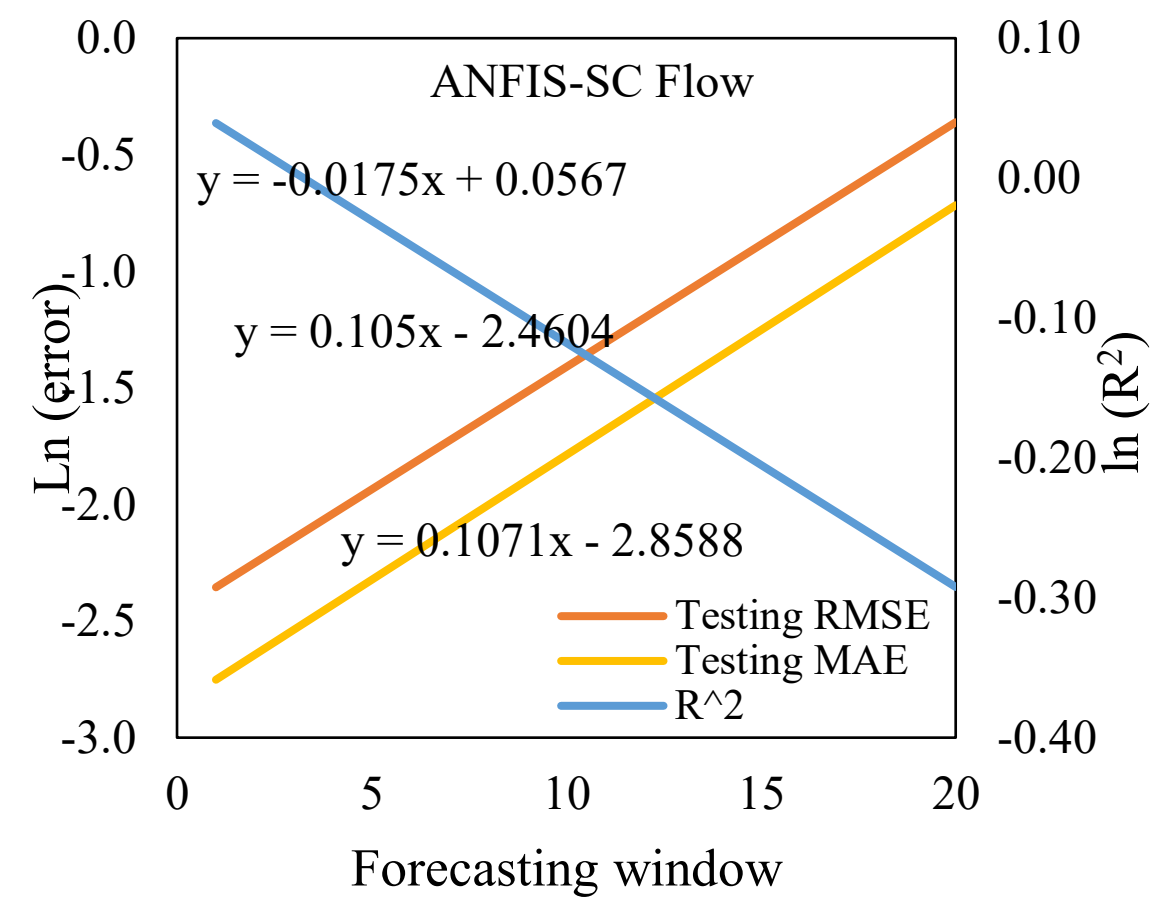

Testing RMSE: $y = 0.105x - 2.4604$
Testing MAE: $y = 0.1071x - 2.8588$
$R^2$: $y = -0.0175x + 0.0567$

**ANFIS-FCMC**

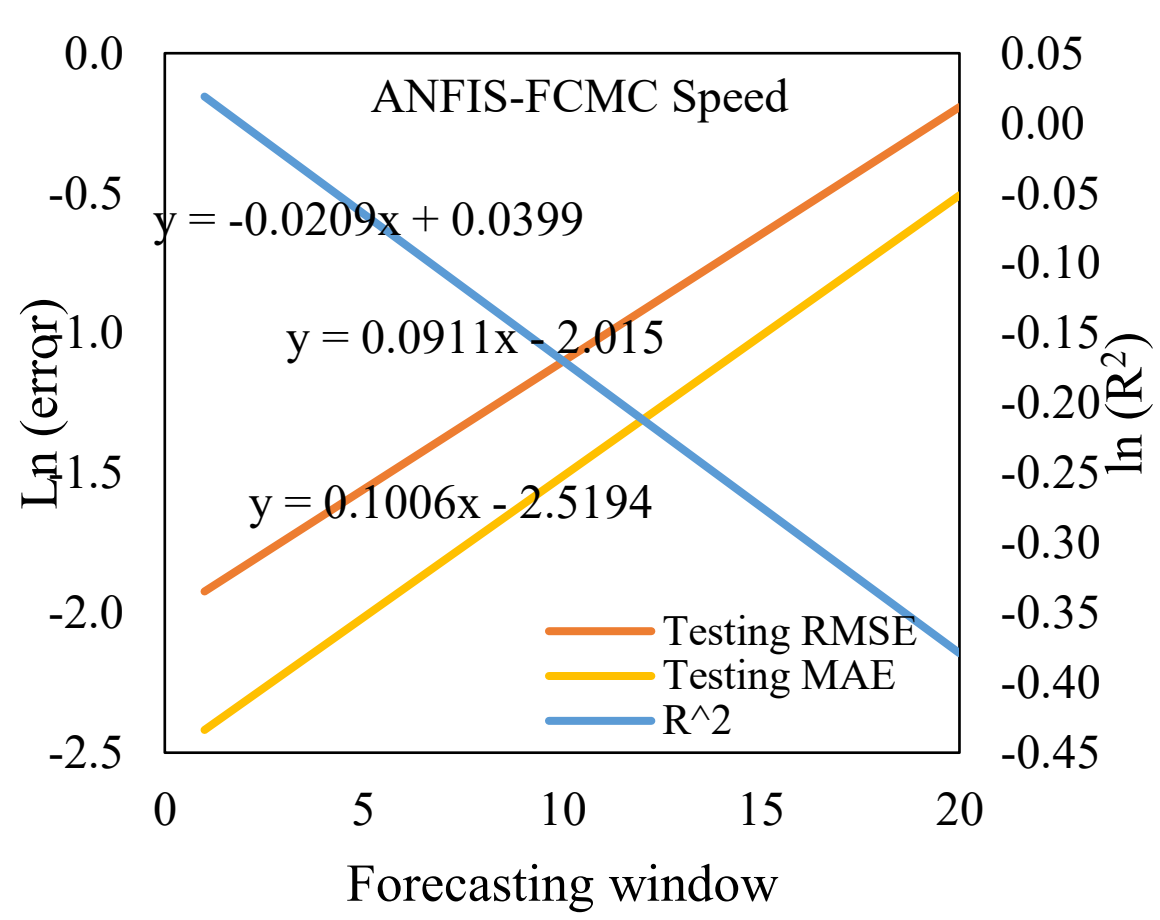

Testing RMSE: $y = 0.0911x - 2.015$
Testing MAE: $y = 0.1006x - 2.5194$
$R^2$: $y = -0.0209x + 0.0399$

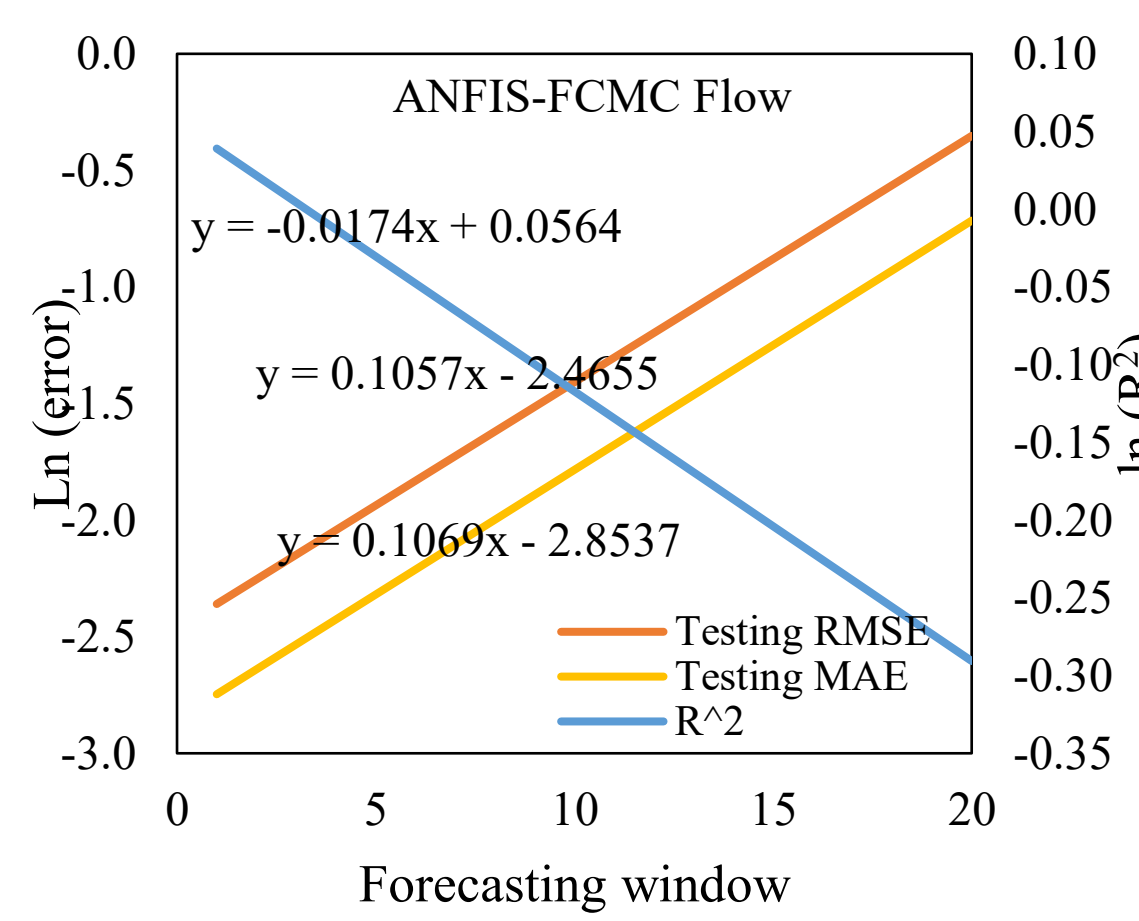

Testing RMSE: $y = 0.1057x - 2.4655$
Testing MAE: $y = 0.1069x - 2.8537$
$R^2$: $y = -0.0174x + 0.0564$

**Model Application**

In this work, various models have been trained to a large dataset of speed and flow data across the roadway segment of Harbor freeway in the first part. The next part of this study involves determining the models representing the performance metrics of the machine and deep learning models for both speed and flow for increasing forecasting windows. A forecasting window of 20 is used in this study which represents 100 minutes because the temporal granularity of the data is 5 minutes. A prediction of traffic flow and speed for 1 hour and 40 minutes is considered to be medium-term traffic flow prediction and is used for various applications such as traffic signal control, ramp metering or even travel time prediction studies (*22,23*). Since it is possible to generate large volumes of data from the loop detectors of PeMS, deep learning methods can be utilized to make such medium term predicts as it shows the most superior performance with this dataset.

## CONCLUSIONS

The results of this study reveal key insights into the predictive capabilities and robustness of various machine learning (ML) and deep learning (DL) models for traffic speed and flow forecasting over increasing prediction horizons. At shorter forecasting windows (e.g., window 1), ANFIS-GP outperforms all models with the lowest RMSE, MAE, and highest $R^2$ values. This is attributed to its grid-partition structure, which allows full coverage of the input space and efficient handling of nonlinear patterns in the data. However, as the forecasting window increases (e.g., windows 10 and 20), Bi-LSTM becomes the top-performing model, maintaining high prediction accuracy due to its ability to capture long-range dependencies in both temporal directions, which ANFIS-GP lacks.

The exponential increase in error with forecasting horizon was confirmed through trendline analysis of RMSE, MAE, and $R^2$ values. By applying natural logarithms, linear relationships were obtained, where the slope indicates the rate of error growth (robustness) and the intercept reflects the initial error. Models with lower slopes—such as Bi-LSTM and Decision Tree Regression—are more robust. In contrast, ANFIS-FCMC exhibits the highest slopes, indicating high sensitivity to time horizon increases. Despite its robustness, Decision Tree Regression has high intercepts, meaning its predictions begin with considerable error, making it less suitable for short-term forecasting.

Furthermore, flow prediction consistently shows higher errors than speed prediction due to greater variability and noise in flow data. In conclusion, ANFIS-GP is best suited for short-term (1 forecasting window), while Bi-LSTM is ideal for longer-term traffic forecasting, with slope-intercept analysis providing a quantitative tool for model selection based on horizon length.

In this study only some the basic ML and DL models have been used to evaluate the performance of the models for predicting traffic speed and flow and its performance with increasing forecasting windows. Recent research has been developing hybrid model architecture that are more efficient that these basic models (*23*). Therefore, the future scope of this research lies in the employment of these hybrid models for performance evaluation at higher forecasting windows.

## ACKNOWLEDGEMENT

The support for this work was provided by the Committee for Advanced Studies Research (CASR), Bangladesh University of Engineering and Technology (BUET) under the resolution number 3(15), 0423042406. Their financial support has immensely helped the completion of this work. AI tool ChatGPT was used for grammar and writing standards.

## AUTHOR CONTRIBUTIONS

The authors confirm contribution to the paper as follows: study conception and design: A. Sherfenaz, N. Haque, M. A. Raihan, M. Hadiuzzaman; data collection: A. Sherfenaz, N. Haque; analysis and interpretation of results: A. Sherfenaz, N. Haque, M. A. Raihan; draft manuscript preparation: A. Sherfenaz. N. Haque, P. S. Prova, M. A. Raihan. All authors reviewed the results and approved the final version of the manuscript.